\documentclass[10pt]{article}

\usepackage[preprint]{tmlr}
\let\cite\citep
\usepackage{float}
\usepackage{needspace}

\usepackage[utf8]{inputenc} %
\usepackage[T1]{fontenc}    %
\usepackage[pagebackref=true,colorlinks,citecolor=brown]{hyperref}
\usepackage[font=small]{caption}
\usepackage{url}            %
\usepackage{booktabs}       %
\usepackage{amsfonts}       %
\usepackage{nicefrac}       %
\usepackage{microtype}      %
\usepackage{xcolor}         %

\usepackage{epsfig}
\usepackage{graphicx}
\usepackage{amsmath,amssymb,mathtools}
\usepackage{multirow}
\usepackage{enumitem}
\usepackage{subcaption}
\usepackage{xspace}
\usepackage{etoolbox}
\usepackage{algorithm}
\usepackage{algorithmic}
\AtBeginEnvironment{algorithmic}{}
\usepackage{placeins}
\usepackage{wrapfig}
\usepackage[capitalize]{cleveref}
\crefname{algorithm}{Algorithm}{Algorithms}

\newcommand{\includegraphicsorplaceholder}[2]{%
  \IfFileExists{#1}{\includegraphics[#2]{#1}}{%
    \fbox{\rule{0pt}{2in}\rule{.9\linewidth}{0pt}}%
  }%
}
\newcommand{\eg}{e.g.\@\xspace}

\newcommand{\scene}{\mathcal{S}}
\newcommand{\param}{\boldsymbol{\theta}}
\newcommand{\obs}{\mathbf{I}}
\newcommand{\forwardmodel}{\mathcal{R}}

\graphicspath{{figures/}}

\title{Infer3D: Inverse Generative Modeling Enables Robust 3D Reconstruction }

\author{%
  \begin{minipage}{\textwidth}
    \centering
    \normalfont
    \textbf{Amani Kiruga}\textsuperscript{1}\quad
    \textbf{Mark Hamilton}\textsuperscript{2}\quad
    \textbf{Joshua B. Tenenbaum}\textsuperscript{3}\quad
    \textbf{William T. Freeman}\textsuperscript{3}\\[5pt]
    \textbf{Vincent Sitzmann}\textsuperscript{3}\quad
    \textbf{Yilun Du}\textsuperscript{1,\textdagger}\quad
    \textbf{Ayush Tewari}\textsuperscript{4,\textdagger}\\[8pt]
    {\small\itshape
      \textsuperscript{1}Harvard University\qquad
      \textsuperscript{2}Google DeepMind\\[2pt]
      \textsuperscript{3}Massachusetts Institute of Technology\qquad
      \textsuperscript{4}University of Cambridge
    }
  \end{minipage}%
}

\begin{document}

\maketitle
\raggedbottom

\begingroup
\renewcommand{\thefootnote}{\textdagger}
\footnotetext{Equal advising.}
\endgroup

\begin{abstract}
Humans can understand the 3D structure of scenes even from highly unfamiliar or extreme viewpoints. Cognitive science suggests this generalization stems from a top-down generative inference strategy. 
In contrast, state-of-the-art 3D reconstruction neural networks typically rely on amortized, feedforward inference. While fast, this pattern-matching approach degrades when faced with out-of-distribution (OOD) scene conditions unseen during training. Inspired by the generative nature of human vision, we propose Infer3D, an inverse generative modeling framework that recovers the latent variables of a generative model to best match the monocular input. To prevent degenerate flat outputs, we couple the generative image prior with a frozen 3D lifting network that constrains the optimization space to a valid, multi-view-consistent 3D manifold. Across a variety of controlled synthetic and real-world tasks, Infer3D demonstrates superior OOD reconstruction accuracy compared to feedforward networks trained on equal or larger datasets. Furthermore, mirroring how human perception requires more time to process confusing stimuli, we introduce an adaptive inference strategy that expends more test-time compute on ambiguous OOD examples while efficiently routing in-distribution inputs. Project website: \href{https://amanikiruga.github.io/infer3d-site/}{amanikiruga.github.io/infer3d-site}
\end{abstract}

\section{Introduction}
\label{sec:intro}
The human visual system exhibits a remarkable ability to parse the 3D world from 2D projections. Crucially, this capability is not brittle: humans can readily understand the 3D structure of objects and scenes even when viewed from highly unconventional angles~\cite{yildirim2015efficient}, under strange lighting~\cite{adelson2000lightness}, or under different compositional arrangements~\cite{lake2017building}. A central challenge in perception theory is explaining how human vision can be so robust. A rich body of literature in cognitive science and developmental psychology suggests that this robustness arises because human vision is not merely a passive, bottom-up filter but also includes an active, top-down generative process. By
maintaining an internal mental model of the physical world, the mind can
continuously predict and match sensory inputs, allowing it to reliably resolve
ambiguities in unseen environments~\cite{spagna2024visual, clark2013whatever, friston2010free,
tenenbaum2011how}.

This cognitive perspective heavily influenced classical computational theories,
which formalized perception as ``vision as inverse graphics''~\cite{yuille2006vision}.
In this paradigm, visual understanding is cast as Bayesian inference over a
generative model of image formation and scene
structure~\cite{knill1996perception, kersten2004object}. Rather than simply
recognizing 2D patterns, the visual system performs
analysis-by-synthesis, actively searching for the underlying 3D scene
representation that best explains the observed 2D image.

\begin{figure}[t]
    \centering

    \begin{minipage}[t]{0.45\linewidth}
        \centering
        \includegraphics[width=\linewidth]{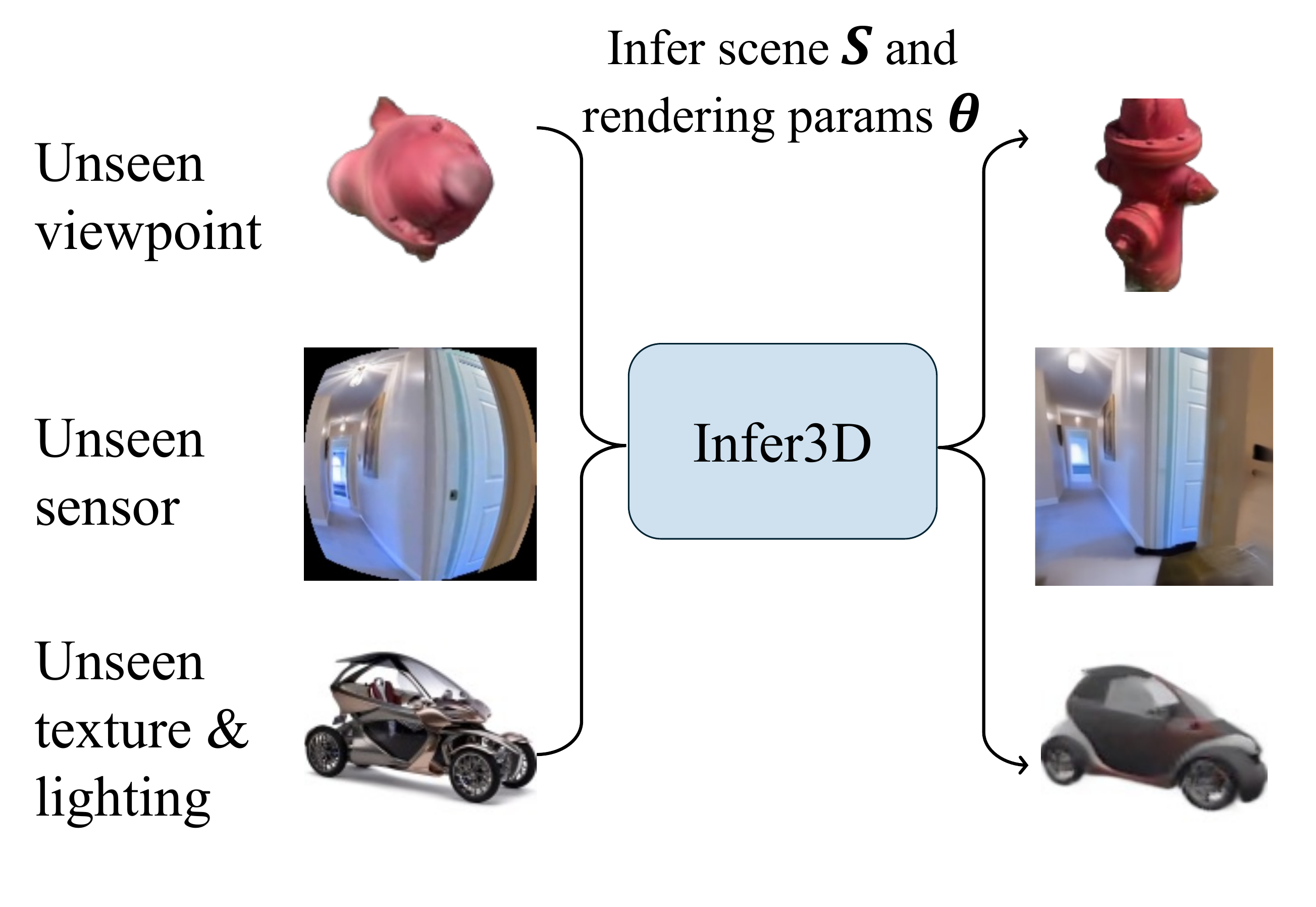}
        \captionof{figure}{\textbf{Robust 3D reconstruction under unseen conditions.} Out-of-distribution (OOD) inputs categorized by domain shifts:  unseen camera viewpoints (top), unseen sensors (middle), and real-world texture and lighting (bottom). {Infer3D} infers the underlying scene ($\boldsymbol{S}$) and rendering parameters ($\boldsymbol{\theta}$) despite these challenging OOD conditions.}
        \label{fig:teaser}
    \end{minipage}
    \hfill
    \begin{minipage}[t]{0.53\linewidth}
        \centering
        \includegraphics[width=\linewidth]{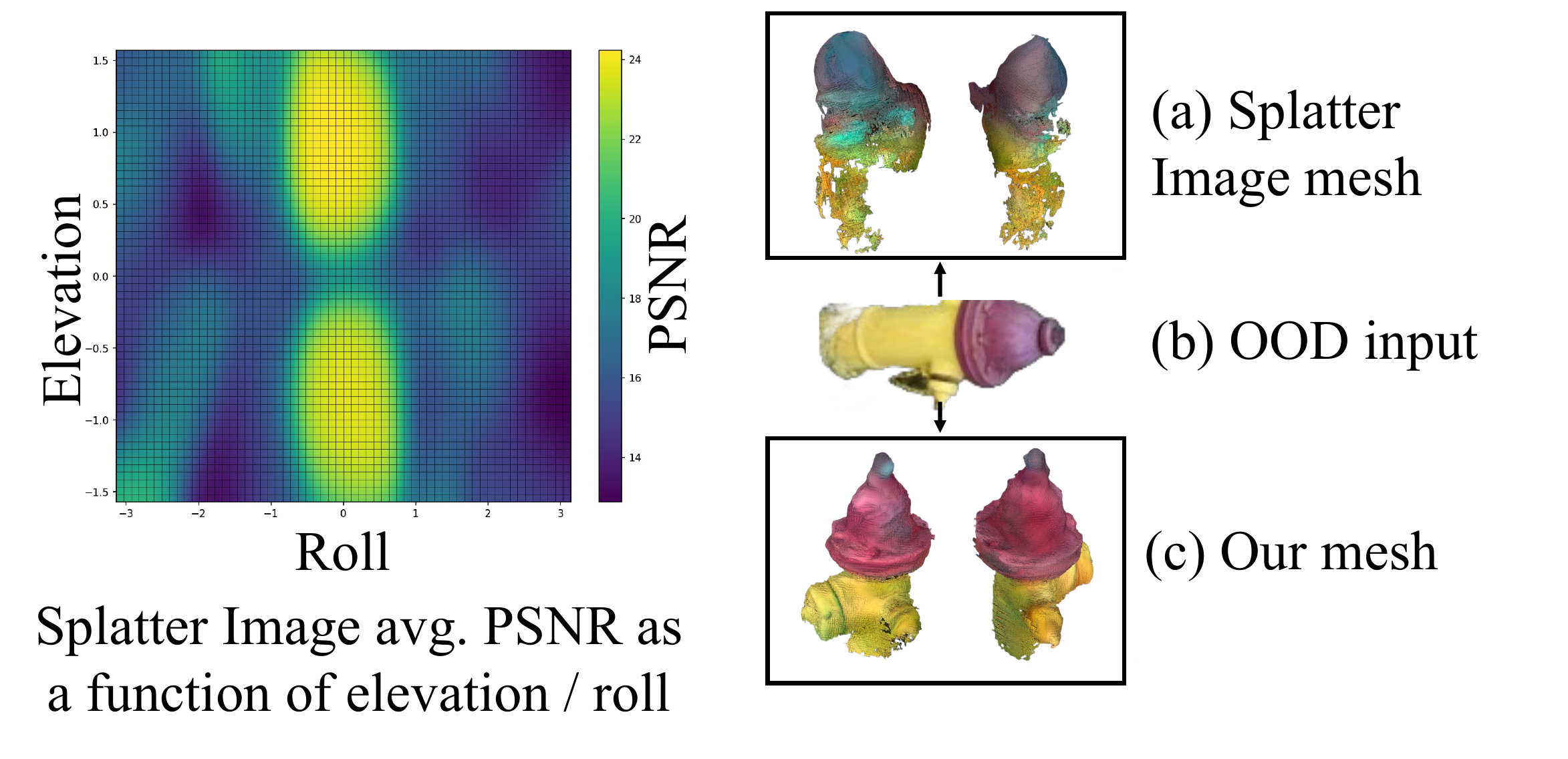}
        \captionof{figure}{\textbf{Failure under Viewpoint Shift.} \textit{Left:} Heatmap of Splatter Image \cite{szymanowicz2024splatter} PSNR evaluated on input at different elevations ($y$-axis) and rolls ($x$-axis). 
        Performance collapses rapidly in regions of OOD elevation or roll. \textit{Right:} Comparison of extracted meshes (Marching Cubes) for an OOD view; the baseline exhibits strong artifacts, whereas our method recovers coherent 3D structure.}
        \label{fig:failure_heatmap}
    \end{minipage}
    \vspace{-20pt}
\end{figure}

In contrast, modern monocular 3D reconstruction systems have largely abandoned
this generative top-down inference in favor of amortized, feedforward architectures~\cite{szymanowicz2024splatter,yu2021pixelnerf,charatan2024pixelsplat,wewer2024latentsplat,hong2023lrm} that directly map a
 2D input image to a 3D output.
While this pattern-matching approach enables very fast inference, it
fundamentally lacks the robustness of generative inference. As we demonstrate
in this paper, feedforward networks degrade in unpredictable ways when faced
with out-of-distribution (OOD) scene conditions unseen during training, such
as unconventional camera viewpoints (see~\cref{fig:failure_heatmap}).

Inspired by the generative nature of human vision and classic analysis-by-synthesis methods~\cite{blanz2023morphable, yuille2006vision}, we propose Infer3D. Rather than trusting a feedforward encoder on OOD data, Infer3D formulates perception as an inverse generative problem. We search for the scene geometry $\scene$ and rendering parameters $\param$ that maximize the explanatory likelihood of the input observation. To ensure our optimization remains constrained to plausible, multi-view-consistent 3D structures, we couple a 2D generative image prior with a frozen 3D lifting network. This encourages the optimized latent variables to always decode to a valid representation on the training manifold, effectively synthesizing an internal hypothesis that best matches the confusing sensory input. Furthermore, mirroring how human perception requires more time to process ambiguous stimuli \cite{kar2019evidence, tang2018recurrent, millidge2024hybrid}, we introduce an adaptive inference strategy. Because our generative framework leverages an encoder for initialization, we show that the initial reconstruction error provides a very reliable and efficient signal for OOD detection. 
This allows Infer3D to dynamically route standard inputs through fast feedforward passes while reserving expensive test-time optimization exclusively for OOD images. 

Overall, our work makes the following contributions: (1) we introduce Infer3D, a modular analysis-by-synthesis framework that combines a learned 2D generative model with a feedforward 3D lifter to enable multi-view-consistent, OOD-robust inference from single images; (2) we propose an adaptive inference and OOD detection strategy that leverages the initial reconstruction error of the generative prior, requiring no external classifiers; and (3) through a controlled experimental setup, we demonstrate that Infer3D successfully generalizes to unseen viewpoints, non-linear lens distortions, and unobserved textures where state-of-the-art feedforward methods collapse (see \cref{fig:teaser}).

\section{Prior Work}

\paragraph{Feedforward Inference. }
There has been a lot of success in models that can reconstruct 3D scenes from monocular images. 
Popular choices of 3D representation are meshes~\cite{liu2019learning,tewari2017mofa,chen2025sam}, or neural radiance fields~\cite{yu2021pixelnerf,tewari2023diffusion,szymanowicz2023viewset, hong2023lrm, xu2023dmv3d}. 
Modern methods rely more on 3D Gaussian representations that also use volume rendering, but are significantly more efficient than neural radiance fields~\cite{charatan2024pixelsplat,szymanowicz2024splatter, xu2024grm, tang2024lgm}. 
In addition, point-based representations are also emerging as a popular choice~\cite{wang2024dust3r,wang2025vggt}. 
These approaches use neural networks to directly transform the input image to the output 3D reconstruction and thus struggle to reason about out-of-distribution settings.

Several papers demonstrate lack of generalization with feedforward inference in out-of-distribution settings~\cite{zhao2024ood,madancapability,alcorn2019strike, abbas2023progress}. 
Mitigating such failures is also a topic of active research. 
Several works have introduced equivariant neural network models that aim to generalize in OOD conditions. 
However, none of these approaches is as general as our method, \eg, we show that our method can be used for images captured under OOD camera poses as well as under OOD textures.
In particular, equivariant approaches either assume 3D or multiview inputs~\cite{xu2022equivariant,deng2021vector,esteves2018learning}, or only include equivariance in parts of their network, \eg, equivariant intermediate representations without equivariant transformations between images and the corresponding representations~\cite{dupont2020equivariant,mitchel2024neural}.
Equivariant models that take 2D image inputs currently only guarantee equivariance under 2D or color transformations~\cite{cohen2016steerable, yang2024learning}.
There is also a line of work in using 3D models for data augmentation~\cite{zhou2023nerf, yang2025novel,zhu2024nerf}. 
However, no such method has been proposed for monocular 3D reconstruction.

\paragraph{Analysis-by-Synthesis.}
Rather than directly transforming an image into the corresponding 3D output with feedforward inference, analysis-by-synthesis solves an optimization problem for inference that searches for the optimal 3D solution that can explain the input observation. 
Because of the ambiguities in the problem, a good prior over 3D models is necessary. 
Many approaches use 3D training data, or pre-captured object geometries, to learn these priors~\cite{blanzoriginal,yuille2006vision,muller2023diffrf, gothoskar20213dp3}. 
Recent papers instead rely on learning priors from 2D data, e.g., using a monocular image generative model as a prior over renderings of the 3D reconstruction~\cite{poole2022dreamfusion,xu2023neurallift,melas2023realfusion,wang2023prolificdreamer}. 
However, monocular priors are not sufficient to enforce multi-view consistency in output. 
Other approaches use multi-view priors~\cite{lin2023magic3d,xu2024grm,long2024wonder3d,szymanowicz2025bolt3d,gao2024cat3d}; however, these methods do not generalize to OOD images as they process input images with neural networks. 
The Bayesian approach of \citet{anciukevicius2022unsupervised} is somewhat related to us, but it makes strong assumptions by assuming known scene parameters for the OOD inputs. 
Recent work on inverse generative modeling for scene understanding~\cite{wang2025compositional} is also closely related in spirit: it formulates perception as inverting a learned generative model and introduces compositional structure to handle complex 2D scenes. 
We are inspired by this view of inference as solving an inverse generative problem, but focus on multi-view-consistent 3D reconstruction and explicit control over camera parameters, rather than 2D compositional layouts.

\paragraph{Hybrid Inference.}
Hybrid inference methods have been explored in the literature~\cite{tewari2018high,alldieck2019learning,leung2022black}. However, they assume that the initial prediction is a reliable initialization—an assumption that breaks down under OOD conditions. In contrast, we propose an adaptive inference strategy that explicitly detects OOD inputs and selectively invokes robust optimization-based inference only when needed.

\paragraph{Optimization-Based Generative Inversion.}
pi-GAN~\cite{chan2021pi} fits the latent conditioning of a 3D-aware GAN to an image but does not infer an unknown camera. BRFI~\cite{pavllo2023shape} and 3D GAN Inversion~\cite{ko20233d}
reconstruct 3D shape by jointly optimizing a generative latent code and
camera pose. Both refine a single initialization obtained from learned
pose or latent predictors, making them sensitive to initialization errors
under distribution shift. FINV~\cite{sun2024partial} filters multiple hypotheses
using provided camera poses and additional observations, whereas Infer3D
searches multiple joint latent and pose hypotheses from a single unposed
image. These methods couple appearance and geometry within a monolithic
3D GAN. Infer3D instead searches a powerful 2D generative prior for
appearance while a separately trained lifter supplies the 3D constraint,
allowing the prior and lifter to be selected independently. We directly compare against BRFI and 3D GAN Inversion in \cref{tab:optimization_baselines_cars,tab:realcars_cd}.

\begin{figure*}[t]
    \centering
    \includegraphicsorplaceholder{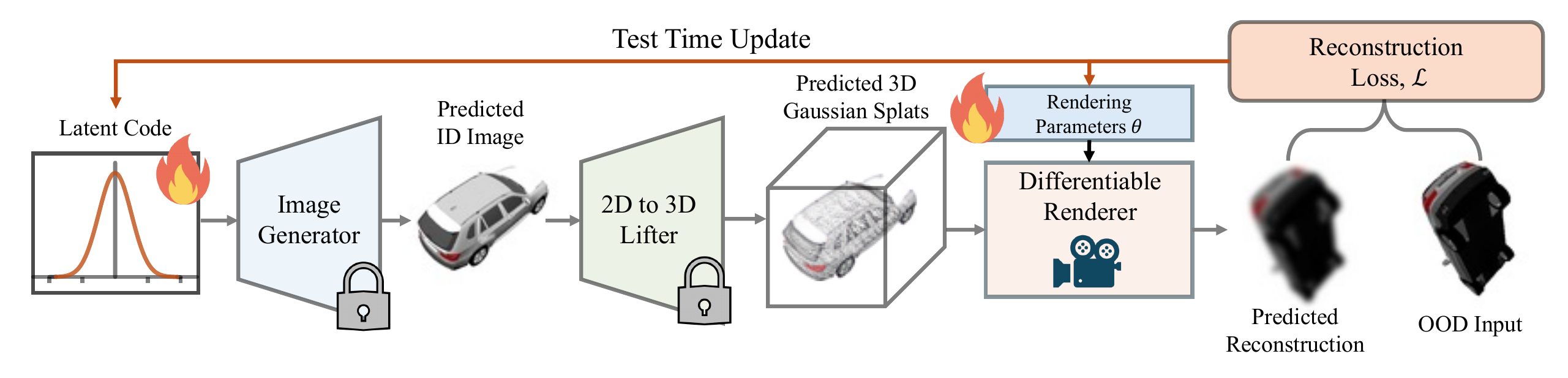}{width=\linewidth}
    \caption{\textbf{Analysis-by-synthesis inference paradigm.} 
    To handle OOD inputs, we employ a test-time optimization loop (red arrow). 
    We freeze the \textbf{Image Generator} and \textbf{2D to 3D Lifter} (indicated by lock icons) to preserve the learned prior. 
    Instead, we optimize the \textbf{Latent Code} and \textbf{Rendering Parameters} (indicated by flame icons) to minimize the reconstruction loss $\mathcal{L}$ between the rendered reconstruction and the OOD input. 
    This ensures the input to the lifter remains a valid predicted ID image while matching the OOD observation.}
    \label{fig:abs_paradigm}
    \vspace{-8pt}
\end{figure*}

\section{Method}
\label{sec:method}

We provide an overview of our method in \cref{fig:abs_paradigm}. 
We first describe the 3D scene representation and our general analysis-by-synthesis framework in \cref{sec:scene_rep}. We then instantiate this framework across three distinct out-of-distribution (OOD) applications in \cref{sec:abs_aplications}: unseen camera poses, unseen sensor distortions, and unseen appearance. Finally, we discuss efficient OOD detection in \cref{sec:ood_detection_method}.

\subsection{Scene Representation and Differentiable Renderer}
\label{sec:scene_rep}
The 3D reconstructions are represented using a scene representation $\scene$ which describes the underlying geometry and appearance of objects in the scene.
This representation can be rendered into an image using a differentiable forward model $\forwardmodel(\scene, \param)$, with rendering parameters $\param$.
Depending on the scenario, $\param$ is parameterized to represent unknown extrinsic camera poses (viewpoint shifts) or expanded to include uncalibrated intrinsic sensor characteristics (e.g., non-linear fisheye lens distortions). 

\paragraph{3D Gaussians.}
In practice, we use 3D Gaussians~\cite{kerbl20233d} as our 3D representation, where the scene is described using a collection of 3D Gaussians that represent its geometry and appearance. 
Each Gaussian is described with parameters that determine its shape, opacity, and color. 
We render the 3D Gaussians using the differentiable splatting-based renderer from 3DGS~\cite {kerbl20233d}, which takes camera parameters as additional input. 

\subsection{Inference via Generative Decoder Inversion}
\label{sec:abs}

\begin{algorithm}[t]
   \caption{Analysis-by-Synthesis via Generative Inversion}
   \label{alg:abs}
   \begin{algorithmic}[1]
      \STATE \textbf{Input:} OOD Image $\mathbf{I}$, Generator $G$, Lifter $\Phi$, Renderer $\mathcal{R}$
      \STATE \textbf{Input:} Schedule $\sigma(t) \rightarrow (k_t, \lambda_t)$, Learning rate $\eta$
      
      \vspace{0.1cm}
      \STATE $\triangleright$ \textit{Initialize latent and camera populations}
      \STATE $\mathcal{Z} \leftarrow \{ \mathbf{z}_r \sim \mathcal{N}(\mathbf{0}, \mathbf{I}) \}_{r=1}^{R}$
      \STATE $\Theta \leftarrow \{ \theta_r \sim \mathcal{U}(\text{SO}(3)) \}_{r=1}^{R}$
      
      \IF{Encoder $E$ is available}
          \STATE $\triangleright$ \textit{Warm-start the first particle}
          \STATE $\mathbf{z}_1 \leftarrow E(\mathbf{I}), \quad \theta_1 \leftarrow \mathbf{I}_{3 \times 3}$
      \ENDIF
      
      \STATE $\bar{\mathbf{z}}_{\text{init}} \leftarrow \frac{1}{|\mathcal{Z}|} \sum \mathbf{z}_r$ \quad $\triangleright$ \textit{Compute regularization anchor}
      
      \vspace{0.1cm}
      \FOR{$t = 1, \dots, N$}
         \STATE $k_t, \lambda_t \leftarrow \sigma(t)$
         
         \STATE $\triangleright$ \textit{Render hypotheses and compute loss for all particles}
         \STATE $\hat{\mathbf{I}}_r \leftarrow \mathcal{R}\big(\Phi(G(\mathbf{z}_r)), \theta_r\big)$
         \STATE $\mathcal{L}_r \leftarrow \mathcal{L}_{\text{recon}}(\hat{\mathbf{I}}_r, \mathbf{I}) + \lambda_t \|\mathbf{z}_r - \bar{\mathbf{z}}_{\text{init}}\|^2$
         
         \STATE $\triangleright$ \textit{Gradient update}
         \STATE $(\mathbf{z}_r, \theta_r) \leftarrow (\mathbf{z}_r, \theta_r) - \eta \nabla \mathcal{L}_r$
         
         \STATE $\triangleright$ \textit{Prune worst-performing particles}
         \STATE Keep the top $k_t$ particles in $(\mathcal{Z}, \Theta)$ with the lowest $\mathcal{L}_r$
      \ENDFOR
      
      \vspace{0.1cm}
      \STATE \textbf{return} Best scene $\mathcal{S}^\star = \Phi(G(\mathbf{z}^\star))$ and pose $\theta^\star$
   \end{algorithmic}
\end{algorithm}

Standard feedforward reconstruction networks function as amortized encoders, mapping observations directly to 3D representations. While efficient, these encoder-based methods lack the inductive biases required for generalization to out-of-distribution (OOD) scene conditions \cite{brady2025generation}. To achieve robustness, we formulate perception as an inverse generative problem via maximum a posteriori (MAP) estimation. Given an OOD observation $\obs_{\text{OOD}}$, we seek the scene parameters $\scene$ and rendering parameters $\param$ that maximize the posterior:
\begin{equation}
    \label{eq:map_posterior}
    p(\scene, \param \mid \obs_{\text{OOD}}) \propto p(\obs_{\text{OOD}} \mid \scene, \param) p(\scene) p(\param).
\end{equation}
Directly modeling a 3D prior $p(\scene)$ is intractable. Instead, we leverage the manifold of a 2D generative model $G(\mathbf{z})$ trained on in-distribution images, composed with a deterministic 2D-to-3D lifter $\Phi: \mathcal{I} \to \mathcal{S}$. This implicitly defines a structural prior where valid scenes are those generated by lifting valid in-distribution images: $\scene = \Phi(G(\mathbf{z}))$, where $\mathbf{z}$ are samples from the latent space of the generative model.
By substituting this parameterization into Eq.~\ref{eq:map_posterior}, the inference reduces to a search over the latent code $\mathbf{z}$ and parameters $\param$:

\begin{equation}
\label{eq:abs_optimization}
\begin{aligned}
\hat{\mathbf{z}}, \hat{\param}
&= \operatorname*{arg\,max}_{\mathbf{z}, \param}
\Big[\underbrace{
\log p(\obs_{\text{OOD}} \mid \scene, \param)
}_{\text{Likelihood}}
\;+\;
\underbrace{
\log p(\mathbf{z})
}_{\text{latent prior}}
\;+\;
\underbrace{
\log p(\param)
}_{\text{rendering prior}}\Big]
\end{aligned}
\end{equation}
In practice, the (unscaled) negative log likelihood is implemented as $\mathcal{L}\Big(
\forwardmodel\big(\Phi(G(\mathbf{z})), \param\big),
\; \obs_{\text{OOD}} \Big)$,
where $\forwardmodel$ is the differentiable renderer from 3DGS and $\mathcal{L}$ is a reconstruction loss. We approximate the prior term $-\log p(\mathbf{z})$ as a quadratic penalty $\|\mathbf{z} - \bar{\mathbf{z}}_{\text{init}}\|_2^2$ centered on the mean of the initial latent population, serving to discourage the optimization from drifting too far from the valid manifold initialization. The final term defines the prior over rendering parameters, described further in Sec.~\ref{sec:abs_aplications}.

\paragraph{Optimization Dynamics.}
We solve Eq.~\ref{eq:abs_optimization} via gradient descent. 
To avoid local minima in the non-convex energy landscape, we employ a multi-start strategy with particle filtering. We initialize a population of latent codes $\{\mathbf{z}_i\}$ and parameters $\{\param_i\}$. The population is iteratively refined, with high-error particles culled periodically, yielding a set of plausible 3D interpretations for the ambiguous OOD input. This procedure is detailed in Algorithm~\ref{alg:abs}.

The formulation in Eq.~\ref{eq:abs_optimization} ensures that the input to the lifting network $\Phi$ is always an in-distribution image $G(\mathbf{z})$, strictly adhering to the manifold where the feedforward model is valid. It is also highly flexible:  we can adapt to distinct domain shifts merely by redefining the optimized rendering parameters $\boldsymbol{\theta}$, their associated prior $p(\boldsymbol{\theta})$, and the likelihood. We instantiate this approach for three specific OOD applications below. 

\subsection{Adapting to Out-of-Distribution Shifts}
\label{sec:abs_aplications}

\paragraph{Out-of-Distribution Camera Poses.} 
For unseen camera viewpoint shifts, we jointly optimize the scene representation alongside the extrinsic camera parameters. Specifically, we optimize the latent code $\mathbf{z}$ and the  camera pose $\boldsymbol{\theta} \in \text{SE}(3)$. We use a uniform prior over the camera pose parameters, continuously projecting rotation matrices onto the $\text{SO}(3)$ manifold via SVD~\cite{levinson2020svd} to ensure stable gradient descent. The likelihood is computed as a combination of MSE and LPIPS losses~\cite{zhang2018unreasonable} to align low-frequency geometry and high-frequency textures. %

\paragraph{Out-of-Distribution Sensor Shifts.} Standard feedforward models assume strict pinhole projection and inherently fail under severe intrinsic parameter shifts, misinterpreting non-linear fisheye lens warping as curved scene geometry. To address this, we optimize the latent code $\mathbf{z}$, camera intrinsics, i.e., focal lengths and principal points ($f_x, f_y, c_x, c_y$), and complex non-linear distortion parameters $\boldsymbol{\phi}$. The distortion prior is parameterized by a Hybrid Neural Distortion Field~\cite{deng2025selfcalibgs}, which is a learnable polynomial fisheye model coupled with an iResNet on a sparse 2D grid. To evaluate the likelihood term, we first warp the perspective-rendered hypothesis generated by our generative model into the fisheye space using the learned forward distortion field ($D_{\boldsymbol{\phi}}$). We then compute MSE and LPIPS losses between this distorted rendering and the input OOD fisheye observation, effectively combining 3D Gaussian Splatting with a hybrid fisheye forward model.

\paragraph{Out-of-Distribution Appearance.} When transitioning from synthetic training data to real-world captures, standard pixel-level reconstruction collapses due to large domain gaps caused by complex reflections, unobserved lighting, and novel materials. 
We can bridge this synthetic-to-real domain gap by leveraging the inherent modularity of our inverse framework.
We replace the pixel-level likelihood in the previous cases with a deep semantic feature loss, maximizing the cosine similarity of DINOv2~\cite{Oquabetal2023} patch-level features extracted from both the OOD input and the rendered 3D hypothesis. 
As semantic features lack fine-grained spatial precision, we couple this with an explicit MSE depth consistency loss to constrain the geometric scale, all rendered via standard pinhole 3D Gaussian Splatting renderer.
Similar to the OOD camera poses setting, we search for the latent code $\mathbf{z}$ and camera pose $\boldsymbol{\theta} \in \text{SE}(3)$.

\subsection{Lifting and Generative Networks}
\paragraph{3D Lifting Network ($\Phi$).}
Our framework is agnostic to the specific architecture of the network $\Phi$. In our experiments, we instantiate $\Phi$ using either:
\begin{itemize}
    \item {Splatter Image}~\cite{szymanowicz2024splatter}, which predicts pixel-aligned 3D Gaussians using a U-Net architecture.
    \item {CATSplat}~\cite{roh2024catsplat}, which utilizes transformer-based context aggregation for more complex geometries.
\end{itemize}
Crucially, $\Phi$ is frozen during inference; we rely on the optimization of $\mathbf{z}$ to adapt the geometry.

\paragraph{Generative Image Priors ($G$).}
We explore two classes of generative models to parameterize the image manifold, see appendix~\cref{fig:priors_detail}. The choice of $G$ dictates the structure of the latent space $\mathcal{Z}$:
\begin{itemize}
    \item {GAN-based:} We employ {StyleGAN2}~\cite{karras2020analyzing}. The latent space is defined as $\mathbf{z} \in \mathcal{W}$, where $\mathcal{W}$ refers to the intermediate 1D representation in the StyleGAN architecture. This generator provides a compact, smooth manifold that is amenable to gradient-based optimization.
    \item {Diffusion-based:} We also utilize {Diffusion Autoencoders (DiffAE)}~\cite{preechakul2022diffusion}. DiffAE encodes an image into a semantic latent $\mathbf{z}_{sem}$ and a stochastic latent $\mathbf{x}_T$. This offers a distinct advantage: unlike GANs, DiffAE includes an explicit encoder $E$ that can initialize $\mathbf{z} = E(\obs)$ for in-distribution data, significantly accelerating convergence (see Sec.~\ref{sec:ood_detection_method}).
\end{itemize}

\subsection{OOD Detection via Generative Prior}
\label{sec:ood_detection_method}
A key capability of our method is that it naturally exposes signals for out-of-distribution (OOD) detection, without requiring the training of additional classifier models.

\paragraph{Inference dynamics as an OOD signal.}
Given an input image, analysis-by-synthesis optimizes latent variables and rendering parameters to minimize a reconstruction loss. For in-distribution inputs $x \in \mathcal{X}_{\text{ID}}$, the generative encoder provides an initialization that lies close to an optimal solution. Consequently, the initial reconstruction error of the generated hypothesis is inherently low. In contrast, OOD inputs $x \in \mathcal{X}_{\text{OOD}}$ lie completely outside the support of the learned image distribution. The encoder fails to provide a near-optimal initialization, resulting in a severe structural mismatch and a high initial reconstruction loss.

We leverage this asymmetry to build an efficient OOD detector. Specifically, we run one forward pass through our method and use the initial reconstruction error (MSE and LPIPS) against the input image as our classification score. If this initial loss exceeds a calibrated threshold, the sample is flagged as OOD and routed to the full optimization loop (Algorithm~\ref{alg:abs}); otherwise, we bypass the optimization and trust the feedforward lifter. 

\paragraph{Automatic Threshold Calibration.}
Importantly, this detection signal requires no additional trained classifiers. To set the decision threshold, we perform an automatic calibration step on a small held-out validation set (5 ID and 5 OOD objects), selecting the threshold that maximizes the F1 score. As we batch all objects and run a single encode-decode pass, this entire one-time calibration process is fast, taking a maximum of 500 milliseconds on CO3D. 
At test time, we can accurately detect OOD inputs in roughly 286 milliseconds without constructing expensive backpropagation graphs. We quantitatively evaluate our fast OOD detection and calibration on CO3D and RE10K~\cite{RealEstate10k} in~\cref{sec:system_analysis}.

\section{Experiments}
\label{sec:experiments}

We evaluate Infer3D's capabilities across three core dimensions: (1) generalization to geometric and camera-level domain shifts, (2) robustness to appearance and environmental domain gaps, and (3) efficacy of our adaptive inference strategy. We report Peak Signal-to-Noise Ratio (PSNR, $\uparrow$), Structural Similarity Index Measure (SSIM, $\uparrow$), and Learned Perceptual Image Patch Similarity (LPIPS, $\downarrow$)~\cite{zhang2018unreasonable} for novel-view fidelity, Chamfer Distance (CD, $\downarrow$) for 3D geometry error, and the Area Under the Receiver Operating Characteristic Curve (AUROC, $\uparrow$) and F1 score ($\uparrow$) for OOD detection.

\subsection{Geometric and Sensor Shift Generalization}
\label{sec:geometric_shift}

We first evaluate how well our approach recovers consistent 3D geometry under extreme geometric perturbations, specifically out-of-distribution camera viewpoints and non-linear lens distortions.
\begin{figure*}[t]
    \centering
    \begin{subfigure}[b]{0.48\textwidth}
        \centering
        \includegraphicsorplaceholder{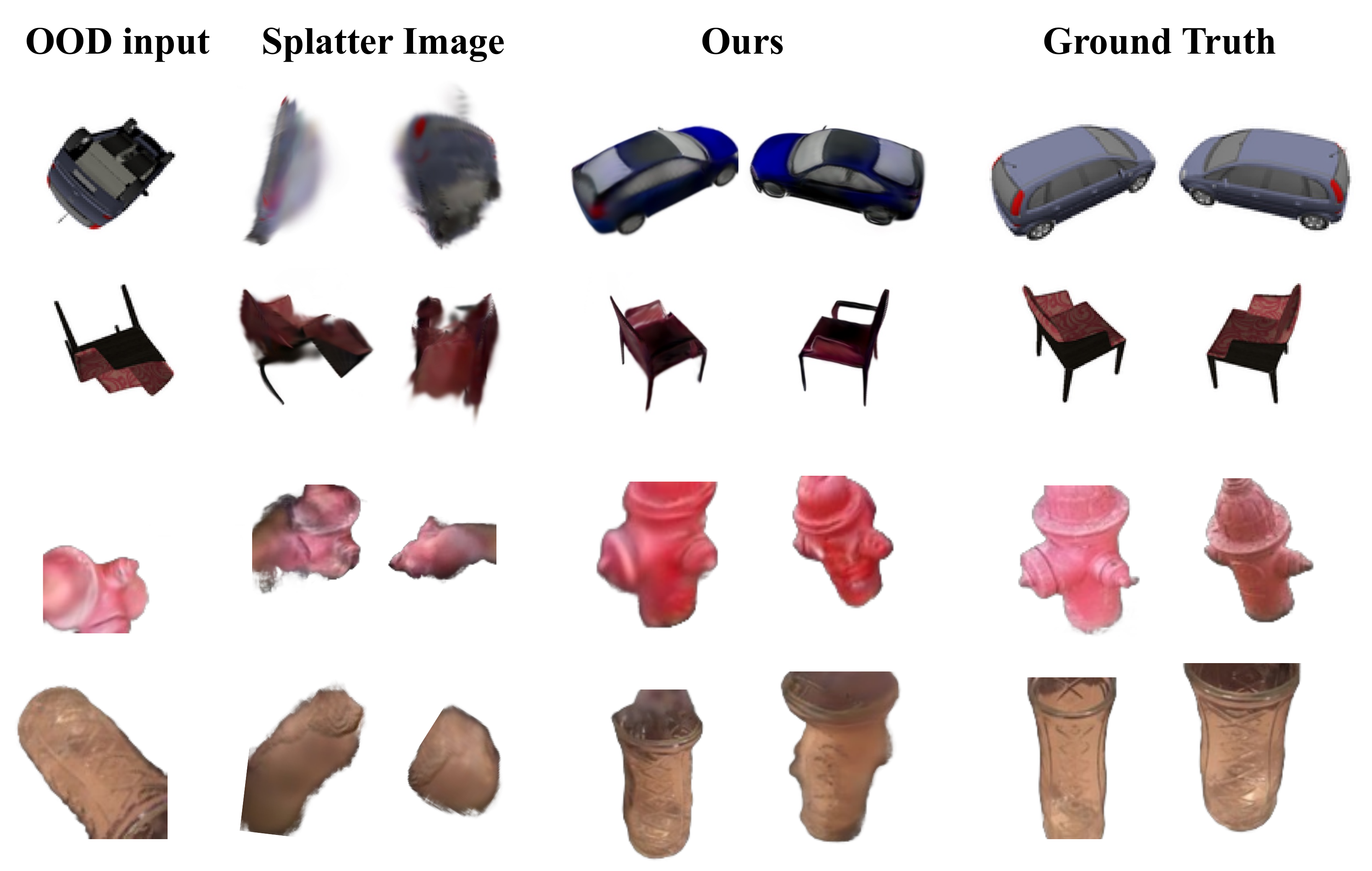}{width=\linewidth}
        \caption{}
        \label{fig:shapenet_results}
    \end{subfigure}
    \hfill
    \begin{subfigure}[b]{0.48\textwidth}
        \centering
        \includegraphicsorplaceholder{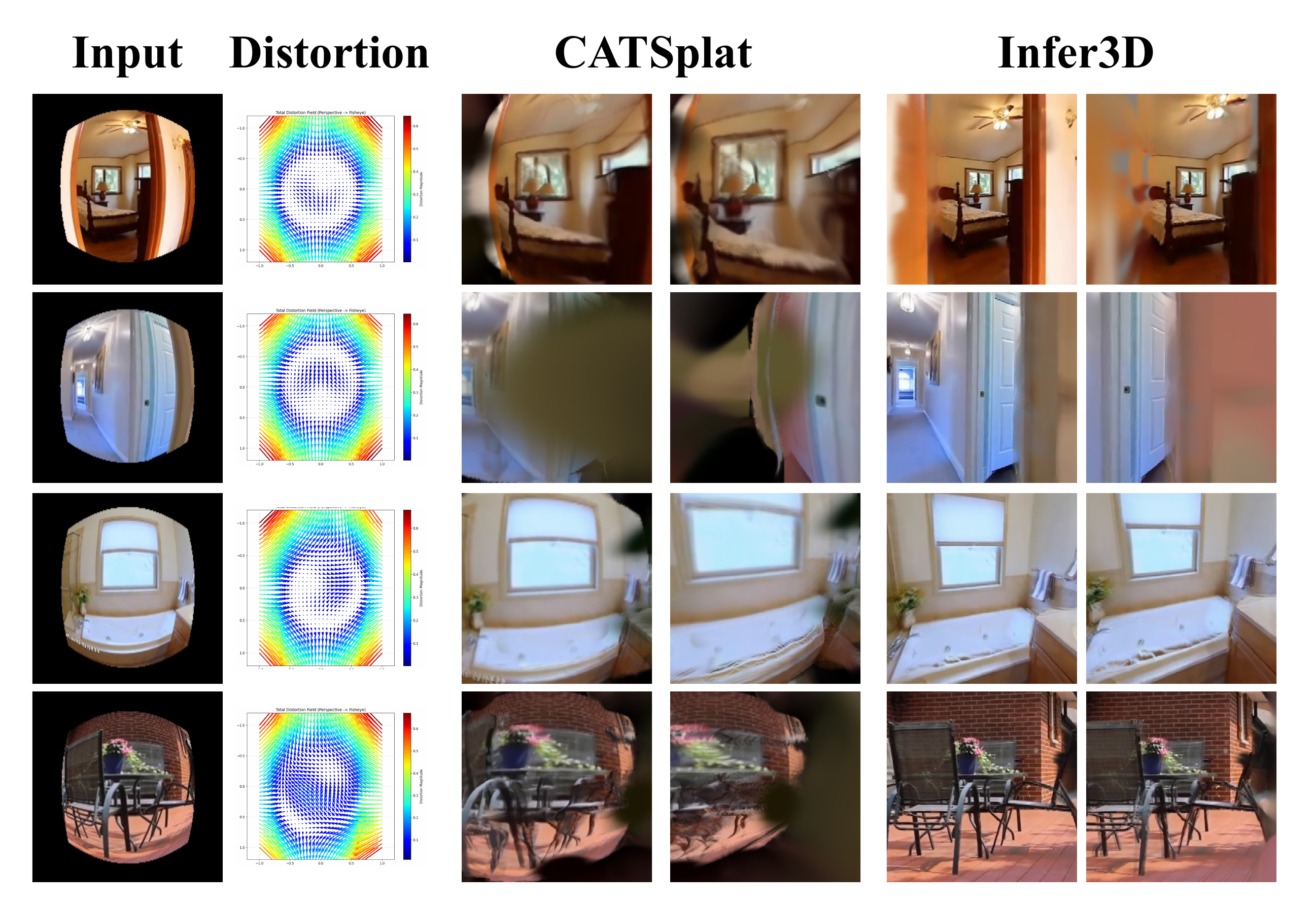}{width=\linewidth}
        \caption{}
        \label{fig:scene_results}
    \end{subfigure}
    
    \caption{\textbf{Robustness to geometric and sensor domain shifts.} 
    \textbf{(a) Viewpoint shift.} Columns show the OOD input, Splatter Image baseline, Infer3D (Ours), and ground truth. Unlike the baseline, our method recovers view-consistent geometry without collapsing or hallucinating structure. 
    \textbf{(b) Sensor shift (fisheye distortion).} Left to right: (1) OOD fisheye input, (2) our learned displacement field, (3) CATSplat baseline (incorrectly modeling distortion as curved geometry), and (4) our disentangled 3D geometry.}
    \label{fig:combined_robustness}
\end{figure*}
\begin{table*}[t]
    \centering
    \caption{\textbf{CO3D Evaluation (Hydrants and Vases).} 
    Metrics: CD $\downarrow$, PSNR $\uparrow$, SSIM $\uparrow$, LPIPS $\downarrow$. 
    \textbf{Bold} = best.}
    \label{tab:co3d_combined}

    \resizebox{\textwidth}{!}{%
        \begin{tabular}{llcccccccc}
            \toprule
            & & \multicolumn{4}{c}{\textbf{In-distribution}} 
            & \multicolumn{4}{c}{\textbf{OOD SE(3)}} \\
            \cmidrule(lr){3-6} \cmidrule(lr){7-10}
            \textbf{Dataset} & \textbf{Method} & 
            \textbf{CD} & \textbf{PSNR} & \textbf{SSIM} & \textbf{LPIPS} & 
            \textbf{CD} & \textbf{PSNR} & \textbf{SSIM} & \textbf{LPIPS} \\
            \midrule
            
            \multirow{5}{*}{\textbf{Hydrants}} 
            & Splatter Image & 
            \textbf{0.052} & \textbf{21.48} & \textbf{0.785} & \textbf{0.168} & 
            0.733 & 15.73 & 0.665 & 0.266 \\

            & Viewset Diffusion & 
            0.145 & 20.95 & 0.724 & 0.216 & 
            0.782 & 14.88 & 0.631 & 0.321 \\

            & CATSplat & 
            0.287 & 18.78 & 0.645 & 0.285 & 
            1.120 & 12.65 & 0.589 & 0.384 \\

            & \textbf{Infer3D (DiffAE)} & 
            0.102 & 20.55 & 0.761 & 0.182 & 
            \textbf{0.461} & \textbf{18.46} & \textbf{0.714} & \textbf{0.220} \\

            & \textbf{Infer3D (StyleGAN)} & 
            0.220 & 17.73 & 0.722 & 0.224 & 
            0.637 & 17.08 & 0.694 & 0.251 \\

            \midrule

            \multirow{5}{*}{\textbf{Vases}}
            & Splatter Image & 
            \textbf{0.061} & \textbf{20.82} & \textbf{0.751} & \textbf{0.179} & 
            0.775 & 15.01 & 0.661 & 0.315 \\

            & Viewset Diffusion & 
            0.138 & 19.98 & 0.684 & 0.245 & 
            0.802 & 14.12 & 0.628 & 0.368 \\

            & CATSplat & 
            0.295 & 17.85 & 0.612 & 0.305 & 
            0.945 & 13.65 & 0.552 & 0.412 \\

            & \textbf{Infer3D (DiffAE)} & 
            0.119 & 19.65 & 0.715 & 0.215 & 
            \textbf{0.552} & \textbf{17.65} & \textbf{0.678} & \textbf{0.261} \\

            & \textbf{Infer3D (StyleGAN)} & 
            0.245 & 16.92 & 0.675 & 0.258 & 
            0.725 & 16.27 & 0.658 & 0.292 \\

            \bottomrule
        \end{tabular}%
    }
    
\end{table*}

\begin{table}
    \centering
    \begin{minipage}[t]{0.48\textwidth}
        \centering
        
        \caption{\textbf{ShapeNet (11 Categories).} Comparison of Splatter Image vs. Infer3D (Ours) (StyleGAN) across difficulty settings.}
        \resizebox{\linewidth}{!}{%
            \begin{tabular}{llcccc}
                \toprule
                \textbf{Setting} & \textbf{Method} & \textbf{CD} & \textbf{PSNR} & \textbf{SSIM} & \textbf{LPIPS} \\
                \midrule
                
                \multirow{2}{*}{\shortstack[l]{\textbf{In-}\\\textbf{dist.}}}
                & Splatter Img. & \textbf{0.122} & \textbf{24.27} & \textbf{0.911} & \textbf{0.107} \\
                & \textbf{Infer3D (SG)} & 0.147 & 21.61 & 0.878 & 0.133 \\
                
                \midrule

                \multirow{2}{*}{\textbf{OOD SO(3)}}
                & Splatter Img. & 0.181 & 18.12 & 0.822 & 0.208 \\
                & \textbf{Infer3D (SG)} & \textbf{0.137} & \textbf{20.67} & \textbf{0.865} & \textbf{0.148} \\
                
                \midrule

                \multirow{2}{*}{\textbf{OOD SE(3)}}
                & Splatter Img. & 0.323 & 14.27 & 0.771 & 0.308 \\
                & \textbf{Infer3D (SG)} & \textbf{0.306} & \textbf{16.30} & \textbf{0.795} & \textbf{0.238} \\
                
                \bottomrule
            \end{tabular}%
        }

        \label{tab:shapenet}
    \end{minipage}%
    \hfill
    \begin{minipage}[t]{0.48\textwidth}
        \centering
        
        \caption{\textbf{RE10K Evaluation.} Comparison of CATSplat vs. Infer3D (DAE) across in-distribution and extreme focal distortion }
        \label{tab:re10k}
        \resizebox{\linewidth}{!}{%
            \begin{tabular}{llccc}
                \toprule
                \textbf{Setting} & \textbf{Method} & \textbf{PSNR} & \textbf{SSIM} & \textbf{LPIPS} \\
                \midrule
                
                \multirow{2}{*}{\shortstack[l]{\textbf{In-}\\\textbf{dist.}}} &
                 CATSplat & \textbf{25.41} & \textbf{0.840} & \textbf{0.149} \\
                & \textbf{Infer3D (DAE)} & 24.03 & 0.801 & 0.186 \\

                \midrule

                \multirow{3}{*}{\shortstack[l]{\textbf{OOD}\\\textbf{Fisheye}}} &
                 CATSplat & 16.25 & 0.610 & 0.327 \\
                & Equidistant & 19.24 & 0.688 & 0.253 \\
                & \textbf{Infer3D (DAE)} & \textbf{23.33} & \textbf{0.775} & \textbf{0.198} \\

                \bottomrule
            \end{tabular}%
        }

    \end{minipage}
\end{table}

\begin{figure*}[t]
    \centering
    \includegraphics[width=\textwidth]{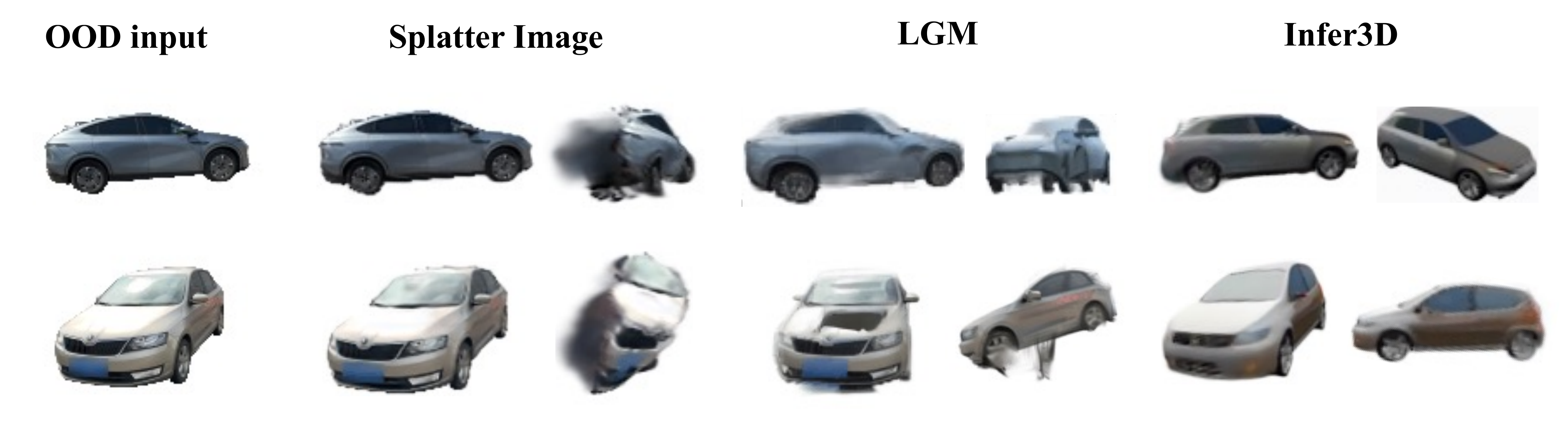}
    \caption{\textbf{Robustness to Unseen Texture and Lighting.} Despite training exclusively on SRN Cars~\cite{sitzmann2019scene}, our method (\textbf{Infer3D}) integrates semantic features (DINOv2) to infer plausible 3D structures unlike feedforward baseline Splatter Image (2nd column) or generative baseline LGM (3rd column) trained on equal or larger (Objaverse) synthetic datasets. }
    \label{fig:unseen_texture_lighting}
\end{figure*}
\begin{table*}[t] 
    \centering
    \begin{minipage}[t]{0.48\textwidth}
        \centering
        \caption{\textbf{Adaptive Inference in Mixed ID/OOD Streams.}
        Performance (PSNR) and amortized runtime using our adaptive
        strategy across simulated real-world distributions on CO3D Hydrants.}
        \label{tab:adaptive_mixed}
        \vspace{-5pt}
        \resizebox{\linewidth}{!}{
            \begin{tabular}{lccccc}
                \toprule
                \shortstack[l]{\textbf{ID/OOD}\\\textbf{Split}}
                & \textbf{F1}
                & \shortstack[c]{\textbf{Splatter}\\\textbf{Image}}
                & \shortstack[c]{\textbf{Opt.}\\\textbf{Only}}
                & \textbf{Adaptive}
                & \shortstack[c]{\textbf{Avg.}\\\textbf{Time}} \\
                \midrule
                99\% / 1\%  & 0.943 & 21.42 & 20.53 & \textbf{21.45} & 2.1s \\
                90\% / 10\% & 0.955 & 20.91 & 20.34 & \textbf{21.16} & 19s \\
                70\% / 30\% & 0.971 & 19.76 & 19.92 & \textbf{20.54} & 55s \\
                50\% / 50\% & 0.986 & 18.61 & 19.50 & \textbf{19.94} & 1.4m \\
                \bottomrule
            \end{tabular}
            
        }
        
    \end{minipage}
    \hfill
    \begin{minipage}[t]{0.48\textwidth}
        \centering
        \caption{\textbf{3D reconstruction accuracy on unseen texture and lighting.}
        Chamfer Distance ($m^2$, $\downarrow$) on ARKit car scenes from
        the RealCars dataset.}
        \label{tab:realcars_cd}
        \resizebox{\linewidth}{!}{
            \begin{tabular}{lcc}
                \toprule
                \textbf{Method}
                & \textbf{Mean CD} $\downarrow$
                & \textbf{Median CD} $\downarrow$ \\
                \midrule
                Splatter Image~\cite{szymanowicz2024splatter}
                & 0.219 & 0.179 \\
                
                Splatter Image + DINOv2 + depth
                & 0.152 & 0.134 \\
                
                LGM~\cite{tang2024lgm}
                & 0.405 & 0.307 \\
                
                SF3D~\cite{boss2024sf3d}
                & 0.242 & 0.206 \\
                
                BRFI
                & 0.491 & 0.502 \\
                
                3D GAN Inversion
                & 0.387 & 0.373 \\
                
                \textbf{Infer3D}
                & \textbf{0.131} & \textbf{0.093} \\
                \bottomrule
            \end{tabular}
        }
        
    \end{minipage}
\end{table*}

\paragraph{Viewpoint Robustness.} 
We evaluate on Multi-Category ShapeNet~\cite{chang2015shapenet} (trained on a fixed $\pm 15^\circ$ elevation ring) and CO3D~\cite{reizenstein2021common} (Hydrants and Vases), see appendix~\cref{fig:dataset_splits}.  We test under random $SO(3)$ rotations and $SE(3)$ translations that severely deviate from the training manifold. We compare against the state-of-the-art feedforward models Splatter Image~\cite{szymanowicz2024splatter}, CATSplat~\cite{roh2024catsplat}, and Viewset Diffusion~\cite{szymanowicz2023viewset}. As shown in Table~\ref{tab:co3d_combined} and Table~\ref{tab:shapenet}, feedforward baselines degrade sharply under OOD shifts, often hallucinating or flattening geometry (Figure~\ref{fig:shapenet_results}). By explicitly inverting generative models, Infer3D successfully matches the observation while maintaining view-consistent structure, recovering performance levels comparable to ID inference.

We further evaluate viewpoint robustness against BRFI~\cite{pavllo2023shape}
and 3D GAN Inversion~\cite{ko20233d} on ShapeNet Cars, for which pretrained generators are available. We use BRFI's released model, which is trained on the full SRN pose distribution, whereas Infer3D's prior and lifter only see restricted elevation rings of ShapeNet Cars. We adapt 3D GAN Inversion using the official EG3D~\cite{chan2022efficient} Cars checkpoint, following its initialization protocol to train latent and pose encoders on SRN training renders with known poses. On cars with OOD $\mathrm{SO}(3)$ viewpoints, Infer3D achieves 20.29 dB, outperforming BRFI (14.30 dB) and 3D GAN Inversion (15.23 dB), even when the baselines receive ground-truth poses (17.70 and 17.37 dB, respectively). Infer3D also preserves more of its ID performance than BRFI, with a drop of 1.14 versus 5.55 dB (\cref{tab:optimization_baselines_cars}).

For CO3D where scale-depth ambiguity is significant, we also incorporate a geometric consistency loss $\mathcal{L}_{\text{depth}}$ to constrain the optimization to plausible absolute depth scales. We minimize the Mean Squared Error (MSE) between the rendered predicted depth map $\hat{\mathbf{D}}$ and the ground-truth depth $\mathbf{D}$ as $\frac{1}{N} \sum_{i=1}^{N} (\hat{\mathbf{D}}_i - \mathbf{D}_i)^2$.
Without extra guidance, the optimization struggles to resolve absolute scale in CO3D (as also reported in related works~\cite{szymanowicz2024splatter,tewari2023diffusion, szymanowicz2023viewset}). 

\begin{table*}[t] 
    \centering
    \begin{minipage}[t]{0.48\textwidth}
        \centering
        \caption{\textbf{Out-of-Distribution Detection AUROC.} We evaluate the ability to distinguish between ID and OOD camera shifts.}
        \label{tab:ood_auroc}
        \resizebox{\linewidth}{!}{%
            \begin{tabular}{lccc}
            \toprule
            \textbf{Method} & \textbf{Hydrants} $\uparrow$ & \textbf{Vases} $\uparrow$ & \textbf{RE10K} $\uparrow$ \\
            \midrule
            Infer3D  & \textbf{0.972} & \textbf{0.961} & \textbf{0.997} \\
            Image alignment              & 0.554          & 0.621          & 0.748 \\
            \bottomrule
            \end{tabular}%
        }
        
    \end{minipage}
    \hfill
    \begin{minipage}[t]{0.48\textwidth}
        \centering
        \caption{\textbf{Calibration Sensitivity.} 
        The detection threshold is highly stable and requires very few validation objects to calibrate (CO3D Hydrants).}
        \label{tab:calibration_sensitivity}
        \resizebox{\linewidth}{!}{%
            \begin{tabular}{lcc}
                \toprule
                \textbf{Num Val Objects} & \textbf{Threshold (ID \%)} & \textbf{F1 Score} \\
                \midrule
                10 & 97.7 & 0.965 \\
                20 & 97.1 & 0.944 \\
                30 & 97.3 & 0.955 \\
                \bottomrule
            \end{tabular}%
        }
        
    \end{minipage}
\end{table*}

\begin{table*}[t]
    \centering
    \begin{minipage}[t]{0.48\textwidth}
        \centering
        \caption{\textbf{Ablation of Inference Losses.} Evaluation of our objective components on CO3D Hydrants. Combining perceptual losses with latent and depth regularization is critical for recovering accurate OOD geometry.}
        \label{tab:loss_ablation}
        \small
        \setlength{\tabcolsep}{4pt}
        \resizebox{\linewidth}{!}{%
            \begin{tabular}{lccc}
            \toprule
            \textbf{Loss Function Configuration} & \textbf{PSNR} $\uparrow$ & \textbf{LPIPS} $\downarrow$ & \textbf{CD} $\downarrow$ \\
            \midrule
            $\mathcal{L}_{\text{MSE}}$ & 14.85 & 0.482 & 0.954 \\
            $\mathcal{L}_{\text{MSE}} + \mathcal{L}_{\text{LPIPS}}$ & 16.20 & 0.315 & 0.725 \\
            $+ \mathcal{L}_{\text{prior}}$ (Mean Latent Dist. Reg.) & 17.45 & 0.255 & 0.580 \\
            $+ \mathcal{L}_{\text{depth}}$ (Full Objective) & \textbf{18.51} & \textbf{0.218} & \textbf{0.458} \\
            \bottomrule
            \end{tabular}%
        }
        
    \end{minipage}
    \hfill
    \begin{minipage}[t]{0.48\textwidth}
    \centering
    \caption{\textbf{Pose and latent optimization ablation.}
    Evaluation on OOD SO(3) ShapeNet Cars using the StyleGAN2 prior.
    Changes are relative to full Infer3D.}
    \label{tab:pose_latent_ablation}
    \small
    \resizebox{\linewidth}{!}{
        \begin{tabular}{lcc}
            \toprule
            \textbf{Configuration}
            & \textbf{PSNR} $\uparrow$
            & $\boldsymbol{\Delta}$ \textbf{PSNR} \\
            \midrule
            \textbf{Full Infer3D (StyleGAN 2)} & \textbf{20.29} & -- \\
            Without pose optimization & 17.19 & $-3.10$ \\
            Without latent optimization & 18.02 & $-2.27$ \\
            Without both & 16.21 & $-4.08$ \\
            \bottomrule
        \end{tabular}
    }
    
\end{minipage}
\end{table*}

\paragraph{Sensor and Field-of-View Shift.} 
Standard feedforward models assume strict pinhole camera projection. We evaluate robustness to intrinsic parameter shifts by simulating realistic wide-angle and radial/tangential fisheye distortions~\cite{lensfun, deng2025selfcalibgs} on RealEstate10K (RE10K). To handle this, we augment our optimization loop with a Hybrid Neural Distortion Field. As shown in Table~\ref{tab:re10k} and Figure~\ref{fig:scene_results}, the baseline (CATSplat) misinterprets lens warping as scene geometry (e.g., curved walls). As a stronger baseline, \emph{Equidistant} is given privileged access to the ground-truth source and target focal lengths and undistorts the input under an equidistant fisheye model before applying CATSplat. Because this model cannot capture the non-linear distortion in our inputs, residual warping remains. Infer3D explicitly models the forward imaging process, successfully disentangling the non-linear lens characteristics from the 3D geometry to recover a clean structural output.

\subsection{Appearance Shift and In-the-Wild Generalization}
\label{sec:appearance_shift}

\paragraph{Unseen Texture and Lighting.} 
We evaluate on the RealCars dataset~\cite{du2025realcar} (in-the-wild captures with complex reflections), restricting the training of our generative prior exclusively to synthetic SRN Cars~\cite{sitzmann2019scene}. For the 3D lifter, we use official Splatter Image weights pretrained on SRN Cars. Because standard pixel-level reconstruction fails under this synthetic-to-real gap, we integrate DINOv2 semantic feature optimization into Infer3D. 
While Splatter Image collapses (Chamfer Distance: 0.219) and generative baselines like Large Multi-View Gaussian Model (LGM) \cite{tang2024lgm}, trained on a much larger Objaverse \cite{deitke2023objaverse} dataset, hallucinate artifacts, our analysis-by-synthesis bypasses the texture nuisance and reconstructs the correct 3D pose and geometry, achieving a significantly lower mean Chamfer Distance of 0.131 (Table~\ref{tab:realcars_cd}, Figure~\ref{fig:unseen_texture_lighting}). For fairness, we also include a finetuned Splatter Image baseline where we provide it with the same DINO features we use as input (PCA-reduced features of channel-32) and pseudo-ground-truth depth from Depth Anything 3~\cite{lin2025depthanything3}. This improves the robustness of Splatter Image (Mean CD of 0.152 from 0.219) but still significantly underperforms Infer3D, demonstrating that inverse generative modeling remains essential for generalization.

The improvement also holds against optimization-based inversion methods trained on synthetic SRN Cars. Infer3D achieves a mean Chamfer Distance of 0.131, compared with 0.491 for BRFI~\cite{pavllo2023shape} and 0.387 for 3D GAN Inversion~\cite{ko20233d}, reducing error by 73\% and 66\%, respectively (\cref{tab:realcars_cd}).

\paragraph{Zero-Shot In-the-Wild Captures.} 
\begin{wrapfigure}[17]{R}{0.5\textwidth}
    \vspace{-40pt}
    \centering
    \includegraphicsorplaceholder{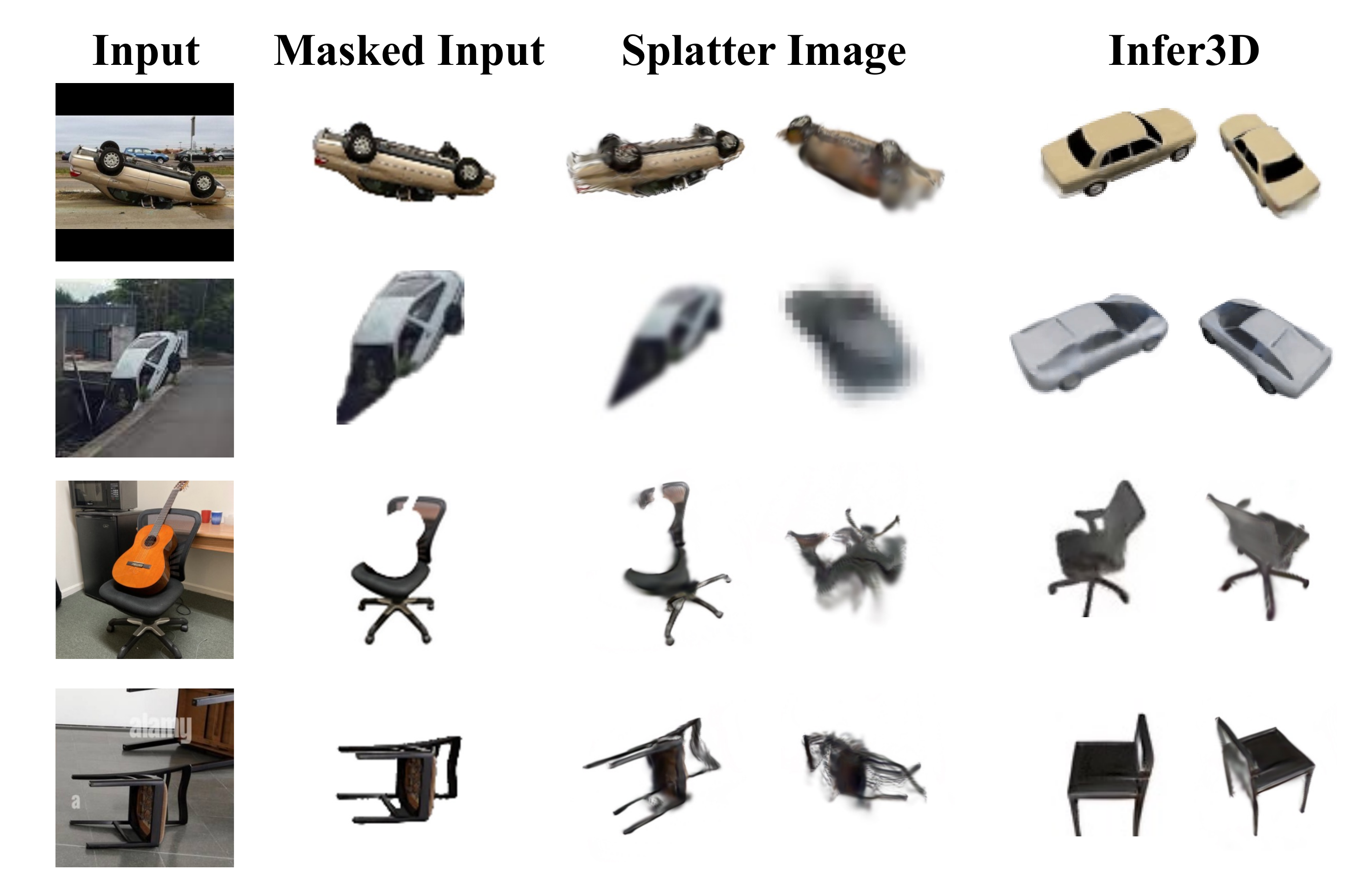}{width=\linewidth}
    \caption{%
    \textbf{Reconstruction from in-the-wild real input images.}
    Each row shows: the original photo; the binary mask fed to the model; the Splatter Image feed-forward prediction; and our analysis-by-synthesis reconstruction. 
    Our method is robust to inputs of varying quality, including low-resolution photos, cluttered backgrounds, and masks that partially occlude the object.}
    \label{fig:real_results}
    \vspace{-40pt}
\end{wrapfigure}
Figure~\ref{fig:real_results} demonstrates Infer3D on casual smartphone photos with varying lighting, partial occlusions, and non-standard cropping. Feedforward baselines fail catastrophically as these images violate strict training preprocessing pipelines. By searching for the camera parameters that best align the generative prior to the noisy image, our compositional approach robustly recovers plausible shapes.

\subsection{Adaptive Inference and System Analysis}
\label{sec:system_analysis}
\begin{figure*}[t]
    \centering
    \begin{minipage}[t]{0.49\textwidth}
        \vspace{0pt}
        \centering
        \small
        \setlength{\tabcolsep}{4pt}
        \resizebox{\linewidth}{!}{%
            \begin{tabular}{lcc}
                \toprule
                \textbf{Method}
                & \shortstack{\textbf{ID SRN Cars}\\\textbf{PSNR} $\uparrow$}
                & \shortstack{\textbf{OOD SO(3)}\\\textbf{PSNR} $\uparrow$} \\
                \midrule
                BRFI
                & 19.85
                & 14.30 (17.70) \\
                3D GAN Inversion
                & 16.22
                & 15.23 (17.37) \\
                \textbf{Infer3D (StyleGAN2)}
                & \textbf{21.43}
                & \textbf{20.29} \\
                \bottomrule
            \end{tabular}%
        }
        \captionof{table}{\textbf{Viewpoint generalization on ShapeNet Cars.}
        We compare our method's performance against other 3D inversion
        methods: BRFI~\cite{pavllo2023shape} and
        3D GAN Inversion~\cite{ko20233d}. All 3 methods are given unposed
        input. Performance with oracle pose is reported in parentheses.
        Oracle-pose variants receive the ground-truth test camera pose.}

        \vspace{-5pt}
        \label{tab:optimization_baselines_cars}
    \end{minipage}
    \hfill
    \begin{minipage}[t]{0.49\textwidth}
        \vspace{0pt}
        \centering
        \includegraphics[width=\linewidth]{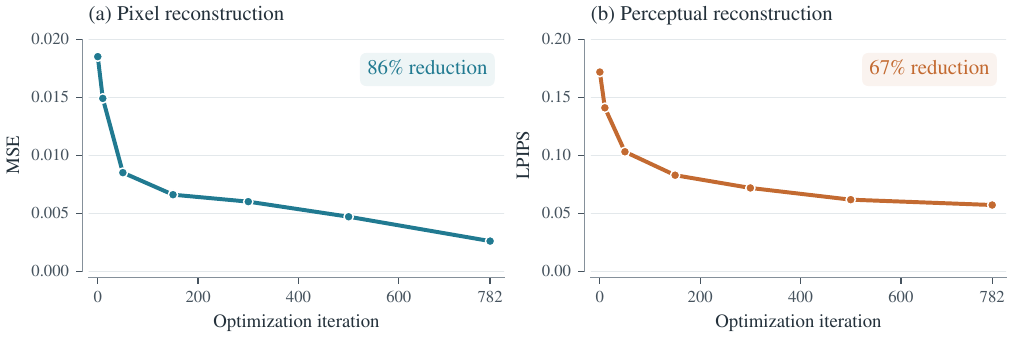}
        \captionof{figure}{\textbf{Optimization convergence on OOD ShapeNet Cars.}
        Unweighted MSE and LPIPS reconstruction terms on OOD ShapeNet
        Cars objects rapidly decrease by 86\% and 67\%, respectively.}
        \label{fig:optimization_convergence}
    \end{minipage}
    \vspace{-10pt}
\end{figure*}
While optimization-based inference is highly robust, it requires more compute than feedforward mapping. We analyze the initial reconstruction error from our generative prior to validate an adaptive system that selectively routes inputs to reduce the effective amortized time per sample.

\paragraph{OOD Detection and Adaptive Routing.} 
We follow the method of \cref{sec:ood_detection_method} to perform OOD detection. We use the encoder of Diffusion Autoencoders (DiffAE)~\cite{preechakul2022diffusion} to provide a near-optimal initialization for ID inputs, which exhibits a low initial reconstruction error. As shown in \cref{tab:ood_auroc}, our initial reconstruction error metric achieves strong detection performance with an AUROC of 0.972 on CO3D Hydrants, 0.961 on Vases, and 0.997 on RE10K. We compare this with an image alignment baseline: the OOD image is fed directly to Splatter Image and the output 3D Gaussians are rendered at identity pose $\mathbf{I}$. We then build a classification score using the reconstruction accuracy of this render and the OOD image. Because encoder-based 3D reconstruction has a bias of copying the image at identity pose, this performs poorly and our method significantly outperforms it while requiring just roughly 286ms of latency. We also apply this adaptive strategy to a mixed stream of ID and OOD inputs (\cref{tab:adaptive_mixed}). In a stream with 10\% OOD data, our system routes standard inputs through the fast feedforward lifter and reserves optimization for outliers, achieving better performance than Splatter Image alone, or optimization alone. The amortized runtime is just 19 seconds per sample. Finally, calibration of our adaptive threshold is sample-efficient and stable across sample sizes. Fitting on merely 10 held-out objects takes $\approx\text{500ms}$ and yields an F1 score of 0.965 (see \cref{tab:calibration_sensitivity}). 

\paragraph{Ablations on Compute Tradeoffs.} 
We run inference for 783 iterations, initializing 600 hypotheses (30 rotations $\times$ 20 latent codes) and pruning to 32, 10, and 5 at iterations 3, 122, and 302. Where available, an encoder provides one warm-start hypothesis. Full optimization settings are given in
\cref{app:optimization_settings}.

We find that the computational cost of Infer3D can be explicitly scaled to prioritize speed without sacrificing robustness. By parallelizing particle optimization ($B=10$), full convergence is achieved efficiently in roughly 2.8 minutes. For applications requiring even faster turnaround, early stopping the optimization loop after just 100 steps yields a rapid 1.2-minute runtime per OOD sample. Notably, this early-stopping configuration still achieves a PSNR of 16.98 which still outperforms the baseline feedforward model Splatter Image (15.73) on OOD CO3D Hydrants. In practice, data streams mix standard inputs with occasional anomalies; pairing rapid OOD detection with analysis-by-synthesis (without early stopping) allows Infer3D to process mixed streams efficiently without bottlenecking pipelines (\cref{tab:adaptive_mixed}). 

\paragraph{Objective and Optimization Ablations.}
Finally, we validate our optimization objective and the necessity of our constrained approach. As detailed in \cref{tab:loss_ablation}, relying solely on pixel-wise MSE for CO3D yields poor geometry. Sequentially integrating perceptual penalties ($\mathcal{L}_{\text{LPIPS}}$), latent regularization ($\mathcal{L}_{\text{prior}}$), and depth consistency ($\mathcal{L}_{\text{depth}}$) is critical for recovering accurate out-of-distribution structures, improving Chamfer Distance from 0.954 to 0.458. Furthermore, we test whether simply updating the weights of the feedforward lifter on the OOD input (trivial test-time training) could achieve similar robustness. \cref{tab:ttt_failure} demonstrates that this naive adaptation collapses the geometry and performs worse than the frozen baseline (12.41 vs. 15.73 PSNR), indicating that optimizing over an internal generative model is more effective than naive test-time adaptation of the lifter.

\looseness=-1
We further examine how pose and latent optimization contribute to our method. On OOD SO(3) ShapeNet Cars, removing pose or latent updates reduces PSNR
by 3.10 and 2.27 dB, respectively, while removing both reduces it by
4.08 dB (\cref{tab:pose_latent_ablation}). We do not change the initialization in these ablations, only whether the target variable is optimized further or selected among initialization candidates. These results support jointly
refining shape and viewpoint during inference and show the gain is spread across the two optimization regimes.

The optimization also progressively improves the fit to the input image,
reducing MSE by 86\% and LPIPS by 67\% over the reported checkpoints
on OOD ShapeNet Cars (\cref{fig:optimization_convergence}).

\section{Limitations and Conclusion}
\label{sec:limitations}

\paragraph{Limitations.} 
While Infer3D significantly improves robustness to OOD shifts, several limitations remain. 
While our method excels at OOD recovery, purely optimization-based inference can result in slightly lower quantitative metrics on in-distribution data compared to a specialized feedforward network. However, our proposed adaptive inference strategy effectively neutralizes this issue by classifying inputs based on convergence dynamics, ensuring that optimization is only triggered when necessary. We further discuss this drop in appendix~\cref{app:indist_drop}. 
Additionally, the visual sharpness of our OOD reconstructions is bound by the capabilities of the underlying 2D generative prior. In this work, we utilized lighter-weight generators (e.g., DiffAE, StyleGAN2) which can sometimes struggle with high-frequency details. We believe that integrating state-of-the-art diffusion models as the generative backbone would resolve this, albeit at a higher computational cost.

\paragraph{Conclusion.}\vspace{-4pt}
We presented a framework for robust monocular 3D reconstruction that treats inference as an inverse generative problem. We demonstrated that standard feedforward networks overfit to the training distribution, failing catastrophically under viewpoint,  sensor, or appearance shifts. By combining a learned 3D lifter with a test-time optimization loop that enforces a 2D generative prior, we recover accurate geometry in scenarios where baselines produce collapsed artifacts. Furthermore, we showed that the dynamics of this optimization process serve as a strong signal for OOD detection, enabling an adaptive system that is both efficient on familiar data and robust to the unexpected. We hope this work encourages a shift from purely amortized inference toward hybrid methods for reliable 3D perception.

\bibliographystyle{tmlr}
\bibliography{references}

\newpage

\appendix
\raggedbottom

\section{Additional Figures}

Here we provide some additional figures that we reference in the main body text. 

\begin{figure}[htbp]
    \centering
    \includegraphicsorplaceholder{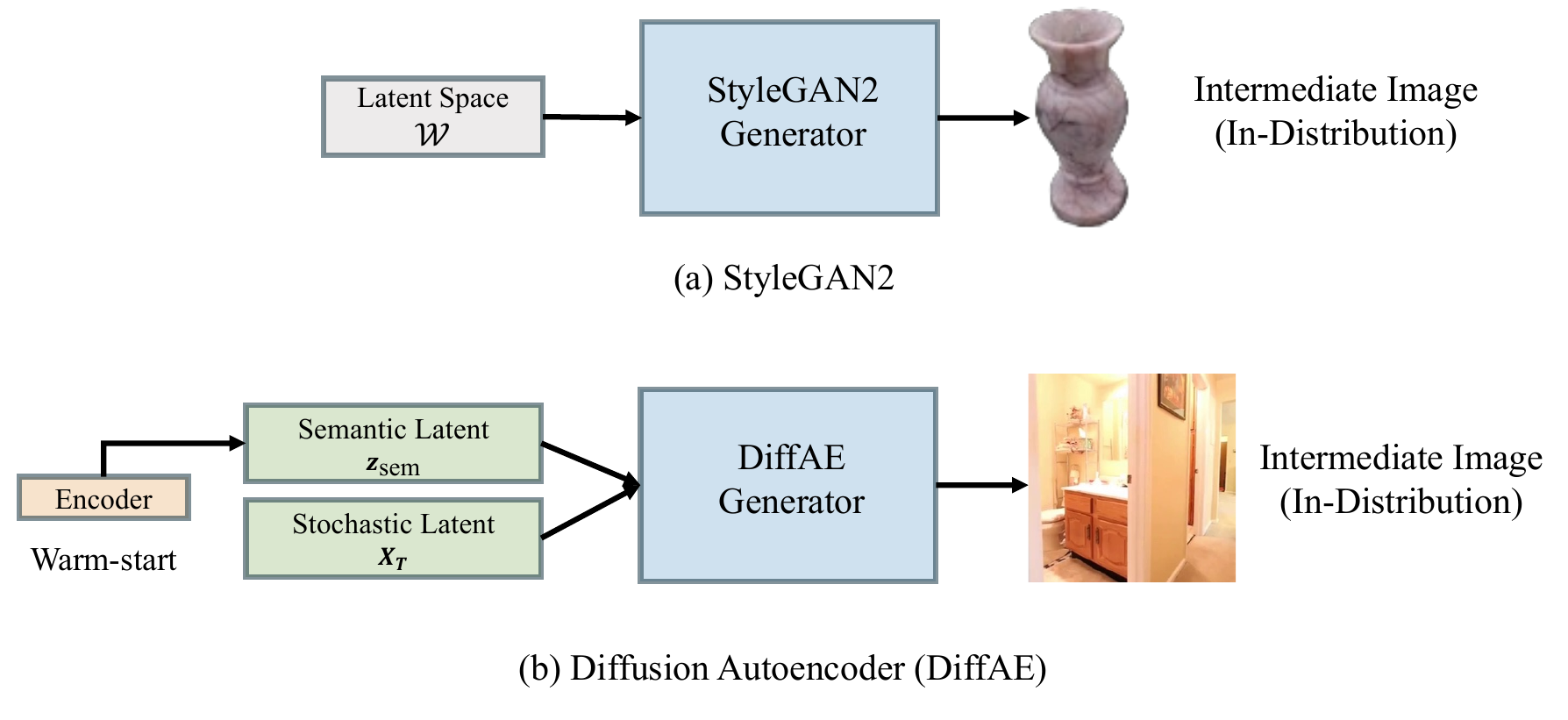}{width=0.7\linewidth}
    \caption{\textbf{Generative Backbone Architectures.} (a) \textbf{StyleGAN2:} Optimization occurs in the $\mathcal{W}$ latent space for semantic consistency. (b) \textbf{Diffusion Autoencoder (DiffAE):} Optimization leverages both a semantic latent $\mathbf{z}_{sem}$ and a stochastic latent $\mathbf{x}_T$. Its architecture allows for an encoder-based warm-start.}
    \label{fig:priors_detail}
\end{figure}

\begin{figure}[htbp]
    \centering
    \includegraphicsorplaceholder{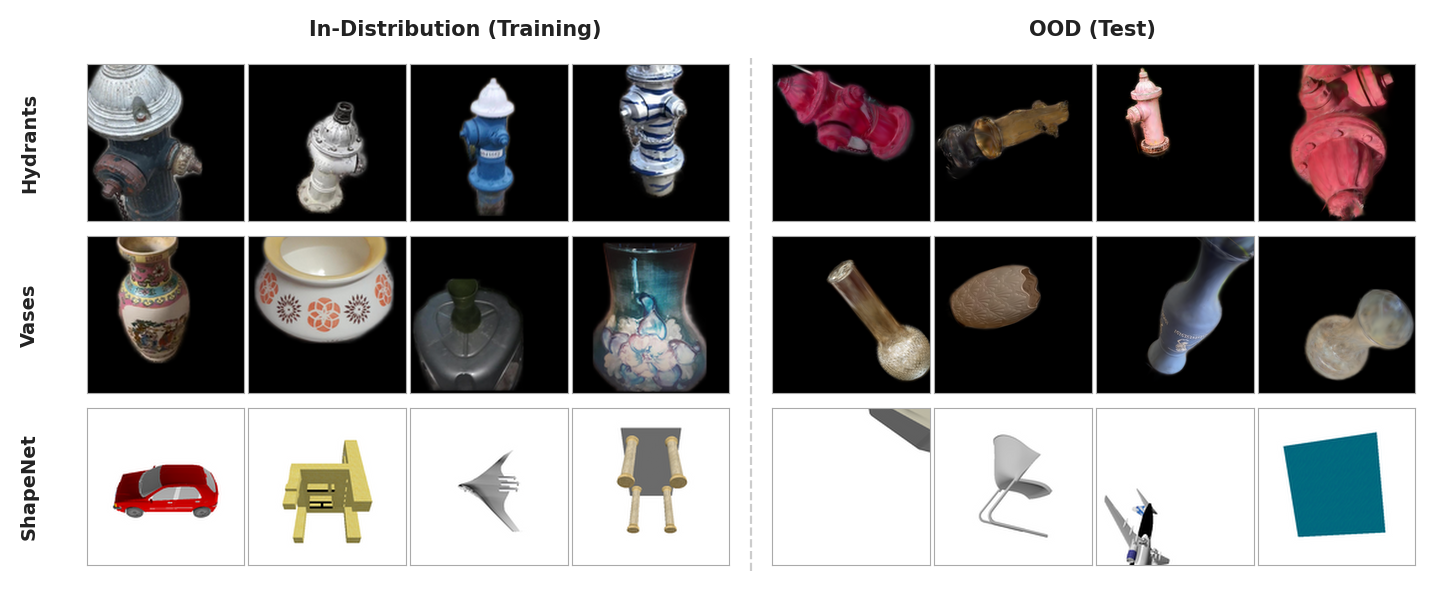}{width=\linewidth}
    \caption{\textbf{Dataset splits for viewpoint robustness evaluation.} \textit{Left:} In-distribution training images from CO3D (Hydrants, Vases) and ShapeNet. These observations are captured from standard, constrained viewpoints, such as fixed elevation rings. \textit{Right:} Out-of-distribution (OOD) test images featuring severe camera pose shifts. These include random $\mathrm{SE}(3)$ translations and $\mathrm{SO}(3)$ rotations over the full viewing sphere, resulting in unconventional angles, extreme elevations, and partial occlusions that challenge standard feedforward priors.}
    \label{fig:dataset_splits}
\end{figure}

\section{Additional Method Details}
\subsection{In-the-wild generalization}
In order to support running our 3D reconstruction inference pipeline on in-the-wild real input images, we first acquire pseudo-GT segmentation masks using Segment-Anything 2 \cite{ravi2024sam2} by initializing multiple point prompts, and selecting the largest mask that is close to the center. We then follow the method of Out-of-Distribution appearance to recover the geometry $\scene$ and camera extrinsics $\param$ (See \cref{sec:abs_aplications}).

\subsection{Compute Resources}
To run all our experiments we use 2 80GB A100 NVIDIA GPUs. At proposal batch size = 10, our full inference occupies a Peak VRAM of 48GB GPU memory. 

\subsection{Optimization Settings}
\label{app:optimization_settings}

Table~\ref{tab:optimization_settings} summarizes the optimization
schedule. Iterations are indexed from 0 to 782. Each iteration updates
all active hypotheses, processing at most $B=10$ simultaneously.
Thus, $B$ controls memory usage rather than the total population size.

\begin{table*}[t]
    \centering
    \caption{\textbf{Optimization schedule.}
    Population sizes, reconstruction-loss weights, and learning rates
    used in the reported inference procedure.}
    \label{tab:optimization_settings}
    \small
    \begin{tabular}{lll}
        \toprule
        \textbf{Parameter} & \textbf{Value or schedule} & \textbf{Meaning} \\
        \midrule
        $R$
        & $30 \times 20 = 600$
        & Initial rotation--latent hypotheses \\

        $N$
        & $783$
        & Total iterations \\

        $B$
        & $10$
        & Maximum simultaneous hypotheses \\

        $k_t$
        & $600 \rightarrow 32 \rightarrow 10 \rightarrow 5$
        & Pruning at $t=3,122,302$ \\

        $\lambda_{\mathrm{MSE}}$
        & $1 \rightarrow 10 \rightarrow 2$
        & Changes at $t=302,732$ \\

        $\lambda_{\mathrm{LPIPS}}$
        & $0.5 \rightarrow 2$
        & Change at $t=302$ \\

        $\lambda_{\mathrm{prior}}$ (DiffAE)
        & $0.001$
        & Constant latent-prior weight \\

        $\eta_z$
        & $0.01$ with ramp-up and ramp-down
        & First $5\%$ and final $25\%$ of iterations \\

        $\eta_\theta$
        & $0.01$
        & Constant rendering-parameter learning rate \\
        \bottomrule
    \end{tabular}
    
\end{table*}

\paragraph{Initialization.}
When an encoder $E$ is available, we replace one random joint
hypothesis with $\mathbf{z}_1=E(\mathbf{I})$ and identity camera pose.
This hypothesis undergoes the same optimization and pruning as the
remaining hypotheses. Without an encoder, all hypotheses are randomly
initialized.

\paragraph{Prior-specific regularization.}
For DiffAE, we apply a quadratic latent penalty centered on the mean
of the encoder-embedded initial latent population, with weight 0.001.
For StyleGAN2, we instead use noise-map regularization to penalize
spatial correlations in the optimized noise maps.

\paragraph{Learning rates and batching.}
The latent learning rate uses a base value of 0.01 with a cosine
ramp-up over the first 5\% of iterations and a ramp-down over the
final 25\%. The rendering-parameter learning rate remains 0.01.
When the active population exceeds $B=10$, we process it in successive
batches, completing an iteration after all active hypotheses have
been updated.

\section{Additional Experiments}

\subsection{Component Ablations}
\label{app:component_ablations}

We evaluate optimization variables, loss terms, and initialization
counts on 25 OOD ShapeNet Cars using the StyleGAN2 prior.
Component-removal experiments use identical examples and initializations.
Unless specified otherwise, all settings follow the full configuration.

\begin{table}[h]
    \centering
    \caption{\textbf{Component ablations on OOD ShapeNet Cars.}
    Changes are relative to full Infer3D. The full configuration uses
    30 pose and 20 latent initializations.}
    \label{tab:component_ablation_full}
    \small
    \begin{tabular}{lcc}
        \toprule
        \textbf{Configuration}
        & \textbf{PSNR} $\uparrow$
        & $\boldsymbol{\Delta}$ \textbf{PSNR} \\
        \midrule
        \textbf{Full Infer3D}
        & \textbf{20.29} & -- \\
        \midrule
        Without pose optimization
        & 17.19 & $-3.10$ \\
        Without latent optimization
        & 18.02 & $-2.27$ \\
        Without both
        & 16.21 & $-4.08$ \\
        \midrule
        Without MSE
        & 19.09 & $-1.20$ \\
        Without LPIPS
        & 19.76 & $-0.53$ \\
        Without noise-map regularization
        & 20.17 & $-0.12$ \\
        \midrule
        Pose initializations: $30 \rightarrow 10$
        & 19.97 & $-0.32$ \\
        Latent initializations: $20 \rightarrow 2$
        & 19.49 & $-0.80$ \\
        \bottomrule
    \end{tabular}
    
\end{table}

Removing MSE or LPIPS reduces PSNR by 1.20 and 0.53 dB, respectively,
supporting their complementary contributions. Reducing the number of
pose initializations from 30 to 10 costs 0.32 dB, while reducing latent
initializations from 20 to 2 costs 0.80 dB. Noise-map regularization
has a smaller effect of 0.12 dB.

\FloatBarrier
\subsection{Comparison with test-time training}

We evaluate direct test-time training (TTT) on OOD CO3D Hydrants by
updating the feedforward lifter's weights to match the input image.
As shown in Table~\ref{tab:ttt_failure}, this update scheme degrades
novel-view reconstruction quality relative to the frozen lifter,
whereas Infer3D improves all three metrics. For this setting, optimizing
the generative latent while keeping the networks frozen is more
effective than directly adapting the lifter's weights.

\begin{table}[htbp]
    \centering
    \caption{\textbf{Comparison with direct test-time training on
    OOD CO3D Hydrants.} Best results are shown in bold.}
    \label{tab:ttt_failure}
    \small
    \setlength{\tabcolsep}{6pt}
    \begin{tabular}{lccc}
        \toprule
        \textbf{Method}
        & \textbf{PSNR} $\uparrow$
        & \textbf{SSIM} $\uparrow$
        & \textbf{LPIPS} $\downarrow$ \\
        \midrule
        Splatter Image     & 15.73 & 0.665 & 0.266 \\
        Test-Time Training & 12.41 & 0.569 & 0.382 \\
        Infer3D (DAE)
        & \textbf{18.46} & \textbf{0.714} & \textbf{0.220} \\
        \bottomrule
    \end{tabular}
\end{table}

\FloatBarrier
\subsection{Generative priors trained on diverse categories}
\label{app:eqm_prior}

\begin{figure}[H]
    \centering
    \includegraphicsorplaceholder{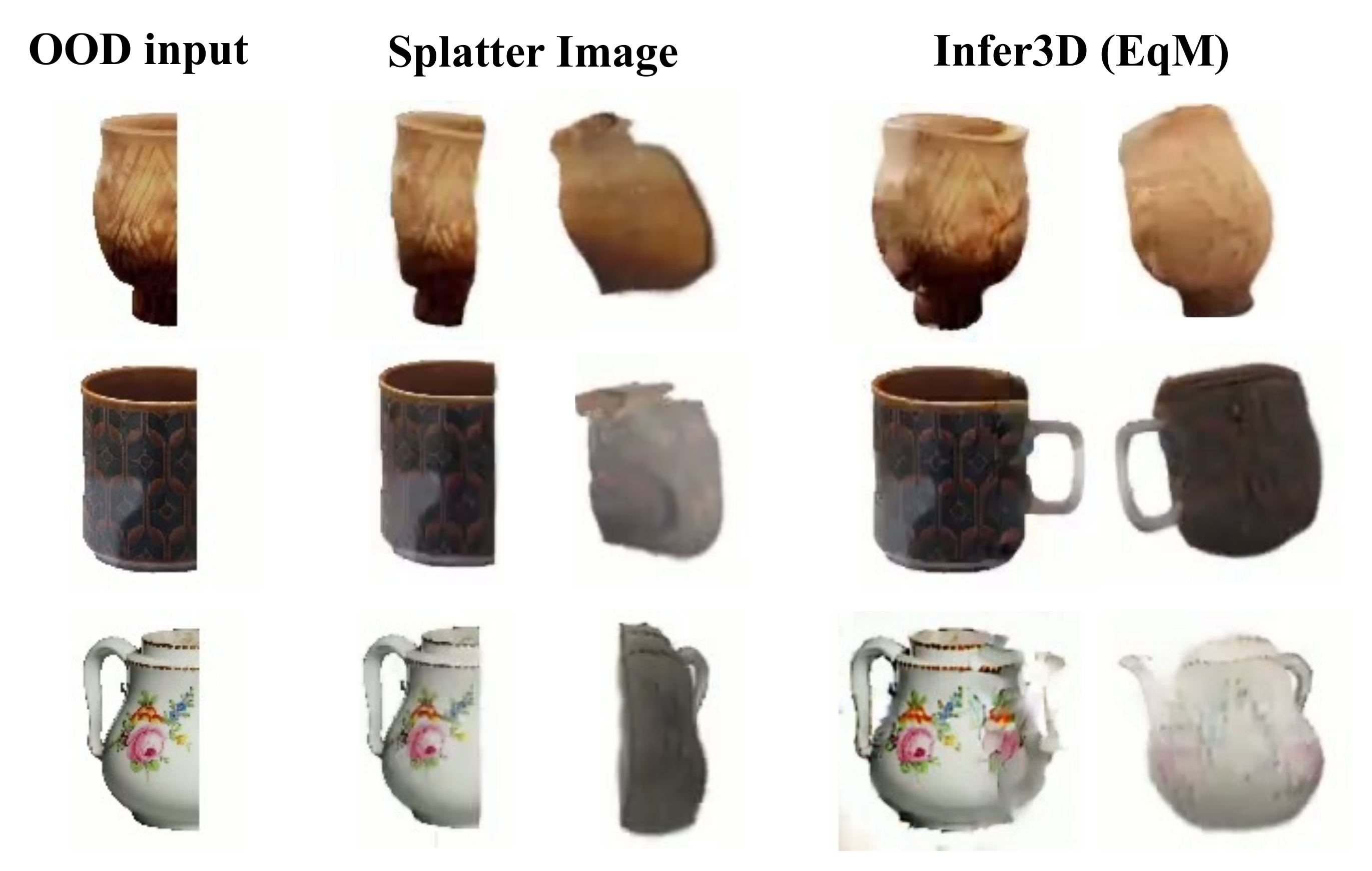}
        {width=0.7\linewidth}
    \caption{\textbf{Qualitative reconstruction with an EqM prior.}
    Left: partially occluded inputs.
    Middle: direct reconstructions from the frozen Splatter Image lifter.
    Right: Infer3D-EqM reconstructions.
    Two views are shown per reconstruction.}
    \label{fig:infer3d_eqm_res}
\end{figure}

To test compatibility with components trained on diverse categories,
we combine a frozen class-conditional Equilibrium Matching (EqM) model~\citep{wang2025equilibrium} trained on ImageNet~\citep{deng2009imagenet} with a frozen Objaverse-trained Splatter Image
lifter~\citep{szymanowicz2024splatter,deitke2023objaverse}. We evaluate on 25 partially occluded objects from a held-out
set spanning six categories, using the same prior--lifter pair
throughout. Infer3D-EqM reduces Chamfer Distance in every category
and by approximately 30\% overall
(Table~\ref{tab:eqm_results}). Figure~\ref{fig:infer3d_eqm_res}
illustrates improvements in object completeness, including the
recovery of missing handles and body regions.

\begin{table}[H]
    \centering
    \caption{\textbf{Reconstruction with an Equilibrium Matching prior.}
    Chamfer Distance (CD) and its relative reduction compared with
    applying the frozen lifter directly.}
    \label{tab:eqm_results}
    \small
    \setlength{\tabcolsep}{6pt}
    \begin{tabular}{lccc}
        \toprule
        \textbf{Category}
        & \textbf{Direct lifter CD} $\downarrow$
        & \textbf{Infer3D-EqM CD} $\downarrow$
        & \textbf{Reduction} \\
        \midrule
        Teapot      & 0.214 & \textbf{0.157} & 27\% \\
        Mug         & 0.225 & \textbf{0.171} & 24\% \\
        Vase        & 0.278 & \textbf{0.180} & 35\% \\
        Pineapple   & 0.147 & \textbf{0.096} & 35\% \\
        Wine bottle & 0.145 & \textbf{0.109} & 25\% \\
        Helmet      & 0.214 & \textbf{0.199} & 7\%  \\
        \midrule
        \textbf{All}
        & 0.206 & \textbf{0.145} & \textbf{30\%} \\
        \bottomrule
    \end{tabular}
\end{table}

\FloatBarrier
\subsection{Comparison with SAM 3D}
\label{app:sam3d}

\begin{figure}[H]
    \centering
    \includegraphicsorplaceholder{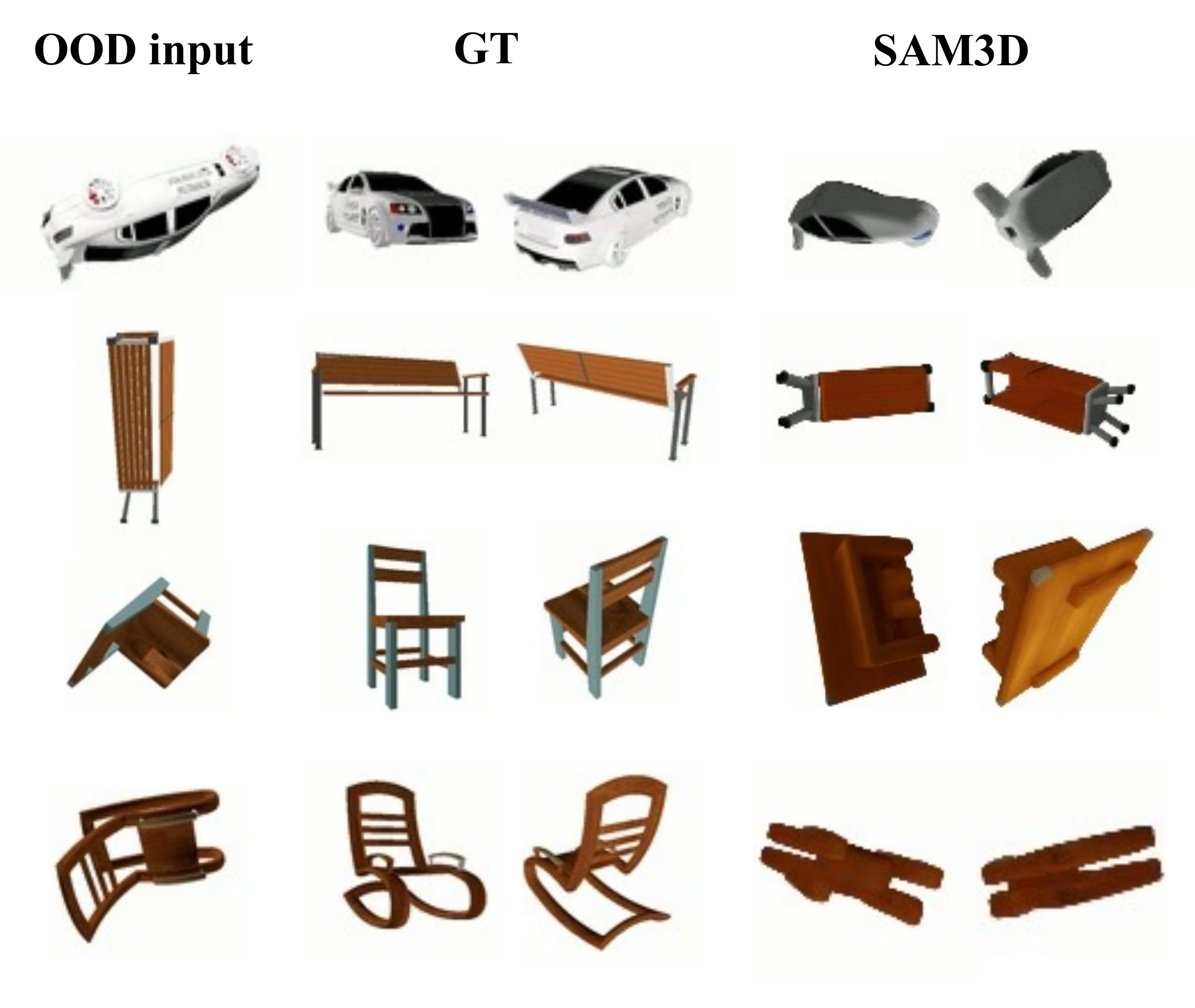}
        {width=0.65\linewidth}
    \caption{\textbf{SAM 3D failure cases under OOD input poses.}
    Each row shows an input image (left), two views of the
    ground-truth shape (middle), and two views of the SAM 3D
    reconstruction (right).}
    \label{fig:sam3d_comparison}
\end{figure}

Table~\ref{tab:sam3d_comparison} compares novel-view synthesis quality
on ShapeNet across 11 categories. Infer3D achieves the best results
on all three metrics under OOD $\mathrm{SO}(3)$ input poses, while
Splatter Image performs best on in-distribution inputs.
Figure~\ref{fig:sam3d_comparison} presents selected SAM 3D failure
cases, illustrating flattened or distorted geometry and missing
structural components under challenging input orientations. We find that even in in-distribution settings, SAM3D does not accurately reconstruct the colors well and so the in-distribution NVS performance is also low. But the shape is not distorted in-distribution as it is OOD. 

\begin{table}[H]
    \centering
    \caption{\textbf{Comparison with SAM 3D on ShapeNet
    (11 categories).} Best results within each setting are shown
    in bold.}
    \label{tab:sam3d_comparison}
    \small
    \setlength{\tabcolsep}{6pt}
    \begin{tabular}{llccc}
        \toprule
        \textbf{Setting}
        & \textbf{Method}
        & \textbf{PSNR} $\uparrow$
        & \textbf{SSIM} $\uparrow$
        & \textbf{LPIPS} $\downarrow$ \\
        \midrule
        In-distribution
        & SAM 3D
        & 16.00 & 0.79 & 0.181 \\
        & Splatter Image
        & \textbf{24.27} & \textbf{0.911} & \textbf{0.107} \\
        & Infer3D (SG)
        & 21.61 & 0.878 & 0.133 \\
        \midrule
        OOD $\mathrm{SO}(3)$
        & SAM 3D
        & 16.18 & 0.79 & 0.194 \\
        & Splatter Image
        & 18.12 & 0.822 & 0.208 \\
        & Infer3D (SG)
        & \textbf{20.67} & \textbf{0.865} & \textbf{0.148} \\
        \bottomrule
    \end{tabular}
\end{table}

\FloatBarrier
\subsection{Potential application to robotic manipulation}

\begin{figure}[H]
    \centering
    \includegraphicsorplaceholder{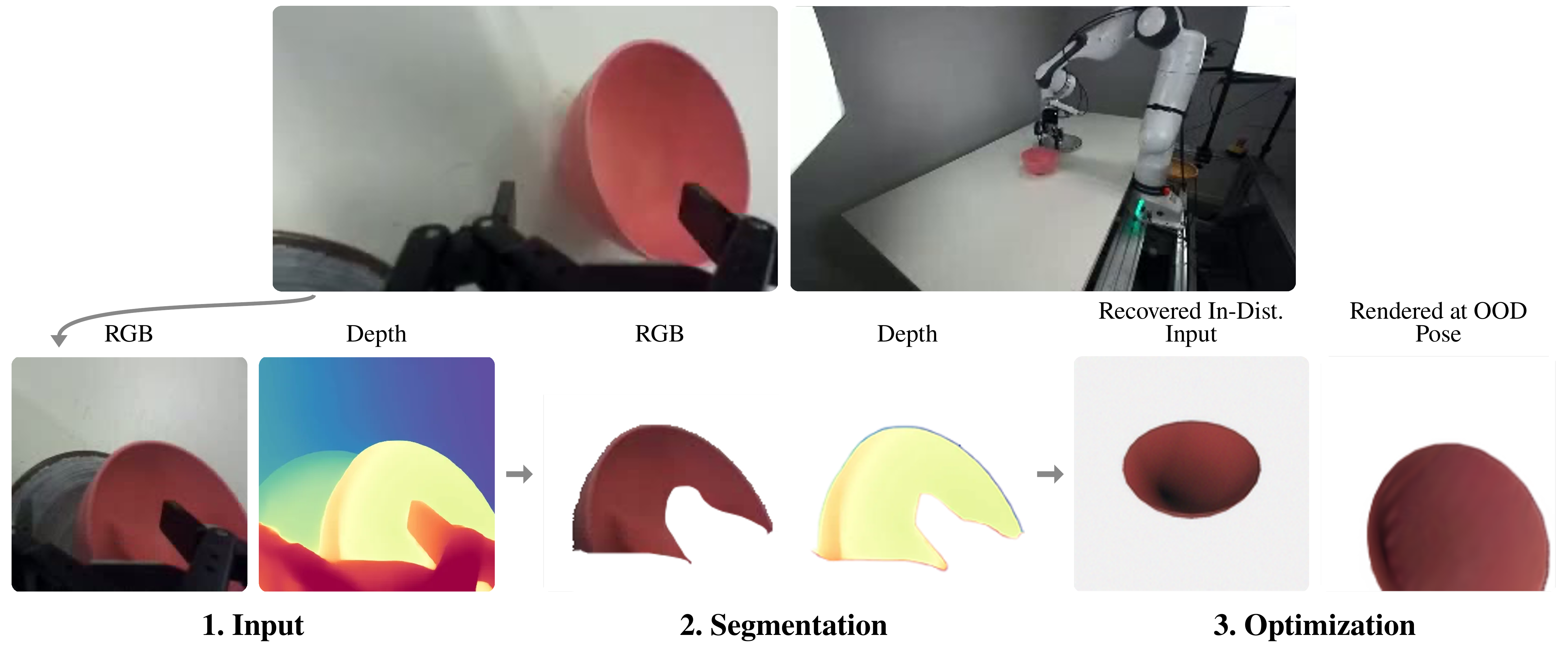}
        {width=\linewidth}
    \caption{\textbf{Reconstruction pipeline for a robotic
    manipulation scene.}
    A wrist-camera observation undergoes depth estimation and
    object segmentation, followed by Infer3D optimization.
    }
    \label{fig:robot_mani_pipeline}
\end{figure}

Figure~\ref{fig:robot_mani_pipeline} illustrates a potential
perception pipeline for robotic manipulation, where wrist-camera
observations can involve unusual viewpoints and gripper occlusions.
We estimate depth using Depth Anything 3~\citep{lin2025depthanything3}
and segment the target object using SAM 3~\citep{carion2025sam3}.
Given the masked RGB image and estimated depth, Infer3D jointly
optimizes the generative latent and camera parameters to recover
the bowl geometry. This example demonstrates reconstruction from
a partially occluded wrist-camera observation; its utility for
downstream manipulation remains to be evaluated.

\subsection{In-Distribution Trade-off}
\label{app:indist_drop}
Run on every input, the optimizer alone is less accurate than the feedforward lifter on ID data: by 0.93~dB on CO3D Hydrants, 2.66~dB on ShapeNet, and 1.38~dB on RE10K (\cref{tab:co3d_combined,tab:shapenet,tab:re10k}). We perform analysis about this drop and find that the performance gap is concentrated near the input view. On ID CO3D Hydrants, the feedforward advantage is 5.7~dB for views close to the input but only 0.9~dB for the most occluded views. Unlike the feedforward lifter, the lifter inside Infer3D never receives the input image directly, so it cannot copy visible details into nearby views; for views that require inferring occluded 3D structure, Infer3D is nearly as accurate. In practice, an image-conditioned feedforward model is preferable for narrow-baseline rendering close to the input view, whereas Infer3D is more robust for wide-baseline views and 3D geometry. The detector-routed system (\cref{tab:adaptive_mixed}) avoids this cost on inputs classified as ID.

\end{document}